\documentclass[10pt,twocolumn,letterpaper]{article}

\usepackage[pagenumbers]{cvpr}
\usepackage{stfloats}

\usepackage{caption}
\DeclareMathSizes{8.5}{8}{6}{5}
\usepackage{multirow}
\usepackage{tabularx}
\usepackage{colortbl}
\usepackage{tikz}
\usepackage{pifont}

\definecolor{clblue}{RGB}{222,234,246}
\definecolor{dsffnorange}{RGB}{252,224,212}
\definecolor{tablehighlight}{HTML}{EAF4D6}
\providecommand{\cc}[1]{\cellcolor{clblue!50}{#1}}

\definecolor{brhasparse}{HTML}{B95757}
\definecolor{brhacoarse}{HTML}{47789E}
\DeclareRobustCommand{\sparseidx}{\textcolor{brhasparse}{\mathcal{I}_i}}
\DeclareRobustCommand{\fineidx}{\textcolor{brhasparse}{f}}
\DeclareRobustCommand{\coarseidx}{\textcolor{brhacoarse}{c}}

\definecolor{cvprblue}{rgb}{0.21,0.49,0.74}
\usepackage[breaklinks,colorlinks,allcolors=cvprblue]{hyperref}

\title{
What Is Redundant at High Resolution? \\Efficient Attention and Feed-Forward Design for All-in-One Image Restoration
}

\newcommand{\authoremail}[1]{\href{mailto:#1}{\texttt{#1}}}
\author{
  {\normalsize Dongheon Lee\textsuperscript{1,}\thanks{This work was done during an internship at NAVER Cloud.}\enspace
  DongJae Lee\textsuperscript{2}\enspace
  Sewhan Chun\textsuperscript{2}\enspace
  Chanseul Cho\textsuperscript{1}\enspace
  Youngwook Kim\textsuperscript{3}\enspace
  Youngmin Ro\textsuperscript{1,}\thanks{Co-corresponding authors.}\enspace
  Yonghyun Jeong\textsuperscript{2,}\footnotemark[2]}\\[3pt]
  {\small \textsuperscript{1}University of Seoul\quad
   \textsuperscript{2}NAVER Cloud\quad \textsuperscript{3}Kookmin University}\\
  {\footnotesize \authoremail{dslisleedh@uos.ac.kr}\quad \authoremail{dongjae.lee@navercorp.com}\quad
   \authoremail{sewhan.chun@navercorp.com}\quad \authoremail{chanseul2001@gmail.com}}\\
  {\footnotesize \authoremail{youngwook@kookmin.ac.kr}\quad \authoremail{youngmin.ro@uos.ac.kr}\quad
   \authoremail{yonghyun.jeong@navercorp.com}}
}
\hypersetup{
  pdftitle={What Is Redundant at High Resolution? Efficient Attention and Feed-Forward Design for All-in-One Image Restoration},
  pdfauthor={Dongheon Lee, DongJae Lee, Sewhan Chun, Chanseul Cho, Youngwook Kim, Youngmin Ro, Yonghyun Jeong}
}

\begin{document}
\addtocontents{toc}{\protect\setcounter{tocdepth}{-1}}
\maketitle
\begin{abstract}
Multi-scale architectures are widely adopted in all-in-one image restoration~(AIO-IR) to capture larger-scale context at low-resolution stages and refine details at high-resolution~(HR) stages.
However, processing spatially large feature maps at HR stages incurs substantial computational costs and memory traffic, making efficient HR processing challenging.
To address this challenge, we propose High-resolution Global-Local Restoration~(HiGLR), an efficient multi-scale AIO-IR network grounded in an analysis of spatial redundancy in self-attention and convolutional feed-forward networks~(ConvFFNs).
Specifically, this analysis identifies three key redundancies: per-query readout at the finest-scale stage, key routing across neighboring queries at intermediate stages, and spatial mixing in the expanded feature spaces of ConvFFNs.
Accordingly, HiGLR replaces costly self-attention at the finest-scale stage with a learned input-independent kernel.
At intermediate stages, we propose a Block-Routed Hybrid Attention, which combines coarse-scale dense attention with fine-scale block-sparse attention over value residuals relative to coarse summaries, and shares routing indices across neighboring queries.
We further introduce a Dual-Space FFN that performs spatial mixing in a compact space and pointwise nonlinear transformation in an expanded space, enabling efficient kernel fusion.
Combining these designs, HiGLR achieves state-of-the-art performance while delivering up to a $4.24\times$ inference speedup over existing methods.
\textup{[\href{https://github.com/dslisleedh/HiGLR}{Code on GitHub}]}
\end{abstract}

\vspace{-0.6cm}
\section{Introduction}
\label{sec:intro}
\vspace{-0.1cm}
Image restoration~(IR) aims to recover high-quality images from degraded observations, including noise, blur, and haze~\cite{DnCNN,MPRNet,DehazeFormer}.
Unlike conventional methods developed for individual degradations, all-in-one image restoration~(AIO-IR) seeks to address diverse degradations using a single unified model~\cite{AirNet,PromptIR,Onerestore,MOCEIR,R2R}.
To restore images under such varied conditions, the model should utilize both broad contextual information and fine-grained spatial cues~\cite{MPRNet, Restormer}.
Accordingly, U-Net~\cite{UNet}-like multi-scale encoder--decoder architectures are widely used in AIO-IR~\cite{UNet,PromptIR,MOCEIR,AnyIR}, integrating large-scale context with fine-grained spatial details across hierarchical high-resolution~(HR) and low-resolution~(LR) stages.

\begin{figure}[t]
  \centering
  \includegraphics[width=0.95\columnwidth]{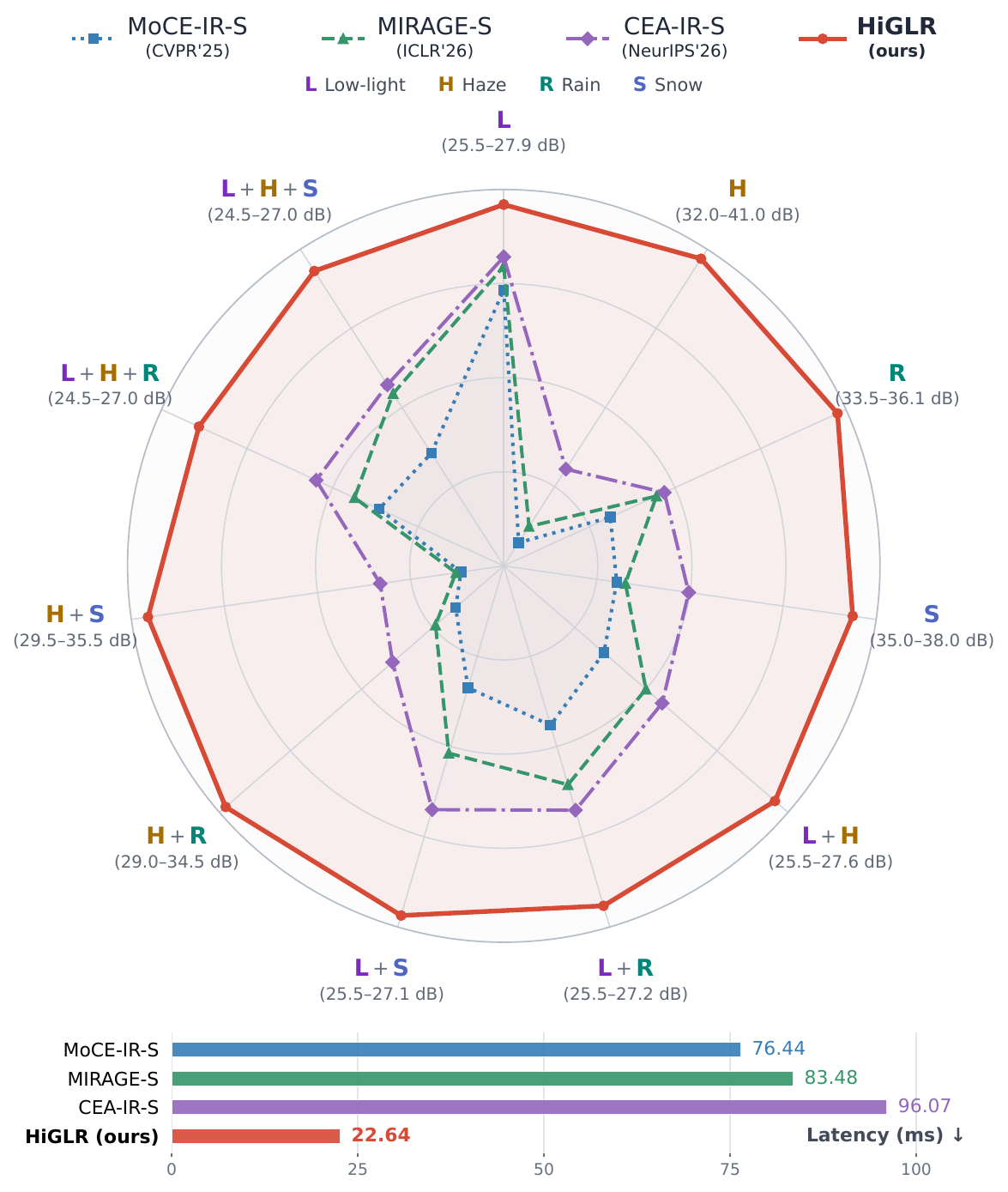}
  \vspace{-0.4cm}
  \caption{
    Comparisons of performance and inference latency on CDD-11.
    HiGLR achieves higher PSNR across all 11 degradation types with substantially lower latency.
  }
  \label{fig:subpixel}
  \vspace{-0.2cm}
\end{figure}

However, HR features are costly to process because of their large spatial dimensions.
For example, spatial self-attention scales quadratically with the number of tokens, making it expensive at HR stages.
While transposed attention~\cite{Restormer} mitigates this quadratic cost, it treats all spatial locations uniformly, hindering spatially-adaptive long-range aggregation.
Meanwhile, although feed-forward networks~(FFNs) scale linearly, materializing their large expanded HR features incurs heavy memory traffic not reflected by MACs.
For a plain FFN, kernel fusion~\cite{FA1, InstantNGP} avoids this materialization, yielding up to a $5.05\times$ inference speedup at identical MACs~(Table~\ref{tab:plain_mlp_fusion_4090}).
Still, widely used Convolutional FFNs~(ConvFFNs)\footnote{In this paper, we refer to Restormer's~\cite{Restormer} GDFN as ConvFFN.} apply depthwise convolution~(DWC) to expanded features, introducing cross-pixel dependencies that complicate such fusion.
Together, the computational and memory overheads of these two core operators pose a substantial challenge to efficient HR processing.
This raises a central question: \textit{Can we substantially reduce the overheads of processing HR features while fully leveraging the benefits of a multi-scale architecture?}

To answer this question, we propose \textbf{High-resolution Global-Local Restoration (HiGLR)}, a multi-scale restoration network guided by stage-wise attention analysis.
At the finest-scale stage, we observe \emph{attention readout redundancy}: the residual attention outputs beyond uniform pooling heavily overlap across spatial queries~(Table~\ref{tab:attention-mass-design}), motivating cheaper spatial mixing with learned, input-independent kernels, as supported by our replacement ablation~(Sec.~\ref{sup_sec:full_resolution}).
Furthermore, at the intermediate resolution stages, attention is more concentrated on a subset of keys, motivating sparse query-key interactions.
Lastly, at the coarsest stage, the smaller token count makes dense attention affordable~\cite{EfficientFormer}.

At intermediate resolution stages, we further observe \emph{routing redundancy}: neighboring queries can share key-selection indices for sparse attention with only a modest reduction in retained attention mass under a small fixed budget~(Table~\ref{tab:attention-mass-design}).
This motivates sharing routing decisions among neighboring queries~\cite{BiFormer,SeerAttention} in block-sparse attention~\cite{SeerAttentionR,HiLSAttention,SAS}, yielding up to a 5.25$\times$ attention speedup at the intermediate stages.
Still, attending to many blocks remains costly, while selecting only a few leaves the remaining regions unaggregated.
We therefore propose Block-Routed Hybrid Attention~(BRHA), which combines dense attention over compressed block summaries with shared-routing sparse attention over value residuals relative to these summaries.
This hybrid design allows sparse attention to focus on a small subset of blocks while maintaining global coverage through compressed block summaries.

Beyond attention, to resolve the HR memory bottleneck in ConvFFNs, we exploit \textit{spatial redundancy} to decouple spatial mixing from nonlinear feature expansion.
Specifically, across two pretrained AIO-IR models~\cite{BioIR,MOCEIR}, we observe that \textit{restoration quality is far less sensitive to output-rank reduction in spatial interaction} than in intra-pixel nonlinearity~(Fig.~\ref{fig:ffn-width}).
Motivated by this asymmetry, we propose the Dual-Space Feed-Forward Network~(DSFFN), which performs spatial mixing in a compact space and injects the resulting spatial context into expanded features for pointwise nonlinear transformation.
Confining cross-pixel dependencies to the compact space keeps the expanded-space operations pointwise, facilitating memory-efficient kernel fusion, especially as the feature expansion ratio increases.

\begin{figure*}[t]
  \centering
  \includegraphics[width=\textwidth]{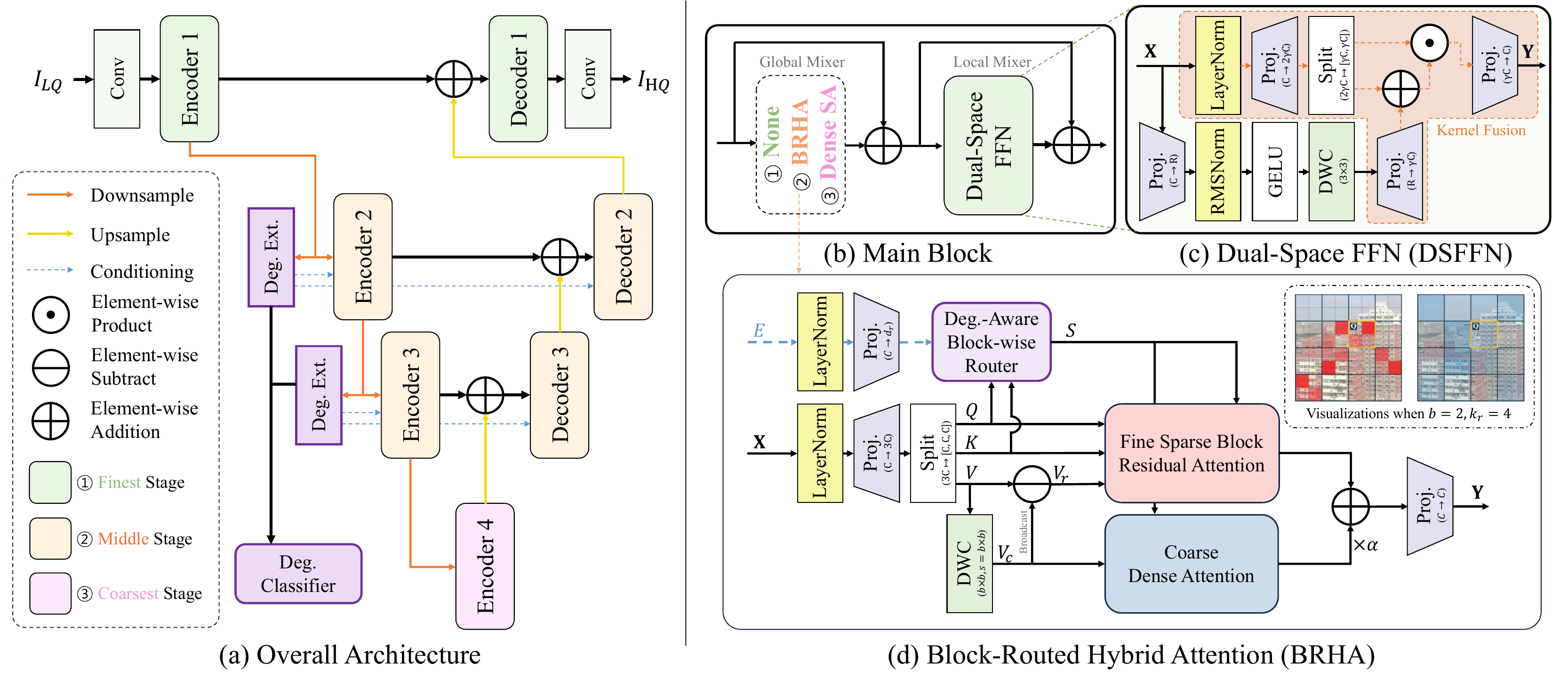}
  \vspace{-0.6cm}
  \caption{
    Overview of HiGLR. (a) The multi-scale architecture and (b) its main block use DSFFN at all scales, BRHA at intermediate resolutions, and dense self-attention at the coarsest scale. (c) DSFFN decouples compact spatial mixing from expanded pointwise transformations. (d) BRHA combines coarse global attention with block-sparse residual attention.
  }
  \label{fig:overall_arch}
  \vspace{-0.5cm}
\end{figure*}

Combining these designs, HiGLR achieves state-of-the-art performance on AIO-3, AIO-5, and CDD-11 with only 7M parameters, while reducing inference latency by 76.4\% and peak GPU memory by 65.8\% compared with CEA-IR-S~(Table~\ref{tab:efficiency}).
Extensive ablations further support our stage-wise design choices and demonstrate the benefits of residual-based hybrid attention and decoupled spatial mixing, while characterizing their quality–efficiency trade-offs.

Overall, our contributions are summarized as follows:

\begin{itemize}
    \item \textbf{Resolution-Aware Architecture via Redundancy Analysis:} We uncover per-query readout, spatial key-routing, and FFN spatial-rank redundancies, and introduce HiGLR to address HR compute and memory bottlenecks without sacrificing multi-scale representations.
    \item \textbf{Block-Routed Hybrid Attention~(BRHA):} We propose an I/O-aware hybrid attention mechanism that shares routing indices across neighboring queries and couples dense attention over block summaries with block-sparse attention over on-the-fly value residuals.
    \item \textbf{Dual-Space Feed-Forward Network~(DSFFN):} We confine depthwise spatial mixing to a compact subspace while keeping expanded nonlinear transformations purely pointwise, enabling end-to-end kernel fusion that drastically reduces memory traffic at HR features.
\end{itemize}

\section{Related Work}
\label{sec:related_work}

\noindent\textbf{Efficient Image Restoration and All-in-One Restoration.}
Efficient restoration backbones~\cite{Restormer,NAFNet} and degradation-adaptive models~\cite{AirNet,PromptIR,MOCEIR,AnyIR,R2R,CEAIR} have advanced the quality--efficiency trade-off in IR and AIO-IR.
For example, Restormer~\cite{Restormer} replaces spatial attention with channel attention to reduce HR costs, while MoCE-IR~\cite{MOCEIR} adapts processing capacity through experts of different complexities.
However, processing large HR feature maps remains costly: spatial attention requires many token interactions, while expanded FFN activations incur substantial memory traffic.
We address these complementary costs through resolution-aware operator allocation, combining compact spatial mixing at full resolution with sparse global aggregation at intermediate stages.

\noindent\textbf{Sparse and Hybrid Attention Methods.}
Sparse self-attention offers a way to reduce the computation and memory costs of dense attention~\cite{Attention} while retaining content-adaptive access to distant tokens.
For example, BiFormer~\cite{BiFormer} routes queries to selected key regions, while SemanIR~\cite{SemanIR} retrieves semantically related keys and shares their indices across layers.
However, aggressive selection limits the spatial information directly available to each query, making global context harder to retain under a small budget.
Hybrid methods address this limitation through compressed-context branches~\cite{NSA} or low-rank attention approximations~\cite{Scatterbrain}.
In contrast, our BRHA exploits the spatial structure of visual features through value residualization: coarse attention aggregates block summaries globally, while sparse fine attention aggregates deviations within selected blocks.

\noindent\textbf{Efficient FFNs and Spatial Mixing Methods.}
ConvFFNs~\cite{Restormer} are widely used in IR to enrich per-token representations with local spatial context~\cite{PromptIR,MOCEIR,BioIR,CEAIR}.
In such designs, materializing expanded HR features incurs substantial memory traffic, while cross-pixel dependencies complicate kernel fusion~\cite{FA2}.
Some prior designs place spatial convolution before channel expansion~\cite{ConvNeXt,FasterNet} or apply it to only a subset of expanded channels~\cite{SMFANet}.
These changes in the placement or channel scope of spatial mixing do not by themselves eliminate the materialization of expanded FFN features.
In contrast, our DSFFN explicitly targets this objective through kernel fusion, confining spatial mixing to a compact space whose width is independent of both the input and expanded widths while keeping expanded nonlinear transformations pointwise.

\section{Proposed Methods}

\subsection{Overall Architecture}
As shown in Fig.~\ref{fig:overall_arch}~(a), HiGLR adopts a four-stage U-Net architecture~\cite{UNet}, following the multi-scale design commonly used in AIO-IR~\cite{PromptIR,MOCEIR}.
An input convolution maps the degraded image $I_{LQ}\in\mathbb{R}^{H\times W\times3}$ to shallow features, which the encoder processes at resolutions from $H\times W$ to $\frac{H}{8}\times\frac{W}{8}$.
Each encoder and decoder stage stacks main blocks (Fig.~\ref{fig:overall_arch}~(b)) that use DSFFN for local mixing and a stage-specific global mixer -- BRHA at the two intermediate scales, dense self-attention at the coarsest scale, and no global mixer at the finest scale.

At the two intermediate scales, degradation extractors map the incoming encoder features to degradation feature maps.
Their block-wise descriptors condition BRHA in both the encoder and the corresponding decoder stage, adapting block scores to the input degradation.
The two degradation feature maps are globally average-pooled and concatenated, then fed to an auxiliary degradation classifier trained with cross-entropy supervision.
The classifier supervises the extractors during training; restoration at inference uses their descriptors without requiring class predictions or degradation labels.

The decoder progressively upsamples features and combines them with same-resolution encoder features through additive skip connections.
Full-resolution refinement aggregates intermediate features through dense connections~\cite{RDN}, and an output convolution produces the restored image $I_{HQ}\in\mathbb{R}^{H\times W\times3}$.

Throughout, $\mathbf{X},\mathbf{Y}\in\mathbb{R}^{H_s\times W_s\times C_s}$ denote a module's input and output at stage $s$, with $\mathbf{Y}$ defined before the enclosing skip connection.
We omit the stage subscript when clear from context.

\subsection{Sparsity and Redundancy in Self-Attention}\label{sec:sparsity}
\noindent\textbf{Stage-dependent attention pattern.}
Self-attention effectively captures long-range dependencies, but its quadratic complexity poses a severe computational bottleneck on HR feature maps.
Sparse attention~\cite{SeerAttention, SeerAttentionR, HiLSAttention, SAS} offers a promising way to reduce this cost, but its suitability depends on how strongly attention mass is concentrated on a small subset of keys.
Notably, as shown in Table~\ref{tab:attention-mass-design}, evaluating a fixed 25\% key-token budget, query-wise key selection retains 90.7\% and 85.6\% of the attention mass at the two intermediate stages, compared with 80.2\% at the coarsest stage and only 58.7\% at full resolution.
This high concentration supports sparse fine-grained aggregation at intermediate resolutions, whereas the reduced token count makes dense attention directly affordable at the coarsest stage.

\noindent\textbf{Per-query readout redundancy at full resolution.}
At full resolution, attention is not only diffuse (58.7\% retained mass), but its residual output beyond uniform pooling is largely shared across spatial queries (92.87\%; Table~\ref{tab:attention-mass-design}).
Similar cross-query redundancy has also been reported in the global context features produced by non-local networks~\cite{GCNet}.
This high readout redundancy indicates limited spatial differentiation on average across the two SA layers, despite costly query-wise aggregation (59.86\,ms).
Together with global context aggregation at lower resolutions, this observation motivates investigating whether lightweight, input-independent local mixing can replace costly full-resolution attention. Our replacement experiment supports this design choice, improving average PSNR by 0.39 dB~(Sec.~\ref{sup_sec:full_resolution}).

\noindent\textbf{Spatial redundancy in key routing.}
Even with concentrated attention at intermediate stages, independently selecting keys for every spatial query incurs substantial routing overhead.
Exploiting spatial correlations among neighboring features, we compare individual and block-shared routing under the same 25\% key-token budget.
Sharing key indices within $8\times8$ query blocks retains 88.0\% and 79.6\% of the attention mass at the two intermediate stages, only 2.7\%p and 6.0\%p below individual routing, respectively.
Neighboring queries can thus share routing decisions with minimal loss in retained attention mass, yielding up to a 5.25$\times$ attention speedup at the intermediate stages, as shown in Table~\ref{tab:stage_attention_latency}.

\begin{table}[t]
    \centering
    \definecolor{evidenceRed}{RGB}{255,65,75}
    \definecolor{evidenceAmber}{RGB}{255,155,0}
    \definecolor{evidenceGreen}{RGB}{20,195,65}
    \newcommand{\evbox}[2]{\begingroup
        \setlength{\fboxsep}{1.5pt}\leavevmode\rlap{\raisebox{0.45ex}[0pt][0pt]{\begin{tikzpicture}[x=1pt,y=1pt,baseline=0pt]
                \path[use as bounding box] (0,0) rectangle (33,0);
                \fill[fill=#1,opacity=.65]
                    (0.0,-3.2) -- (1.3,-1.7) -- (0.2,0.0) -- (1.8,1.2)
                    -- (0.7,3.8) -- (5.5,4.5) -- (12,4.1) -- (19,4.6)
                    -- (26,4.0) -- (32.5,4.4) -- (31.4,2.6) -- (33,1.8)
                    -- (31.8,0.0) -- (32.7,-1.4) -- (31.2,-2.6)
                    -- (32.1,-4.0) -- (25,-4.4) -- (18,-3.9)
                    -- (11,-4.5) -- (5,-4.0) -- cycle;
                \fill[#1]
                    (0.9,-2.1) -- (2.0,-0.8) -- (1.1,1.3) -- (3.0,2.9)
                    -- (10,3.2) -- (17,2.8) -- (24,3.3) -- (32,2.7)
                    -- (30.8,1.1) -- (32.2,-0.2) -- (30.5,-1.1)
                    -- (31.4,-2.7) -- (24,-3.0) -- (17,-2.6)
                    -- (9,-3.2) -- (3,-2.7) -- cycle;
                \draw[white,opacity=.38,line width=.3pt]
                    (3,3.1) -- (12,3.3) (21,-3.1) -- (30,-2.8);
            \end{tikzpicture}}}\phantom{\colorbox{white}{\makebox[30pt]{\strut}}}\llap{\makebox[33pt]{\strut #2}}\endgroup}
    \newcommand{\evlines}[1]{\begin{tabular}[c]{@{}c@{}}#1\end{tabular}}
    \newcommand{\evlabel}[1]{\begin{tabular}[c]{@{}l@{}}#1\end{tabular}}
    \setlength{\fboxsep}{1.5pt}
    \caption{Stage-wise attention statistics and design choices.
Attention statistics are measured on a NAFNet variant with two SA and two FFN sublayers per encoder stage and one NAFBlock per decoder stage.
Retained mass uses a 25\% key budget; block-wise routing shares keys within $8{\times}8$ query blocks.
Readout redundancy is computed by equally averaging the two SA layer-wise energy ratios per stage.
Tints indicate relative assessments within each row.
See Sec.~\ref{sup_sec:stagewise_statistics} for definitions and implementation details.}
    \label{tab:attention-mass-design}
    \footnotesize
    \setlength{\tabcolsep}{0pt}
    \renewcommand{\arraystretch}{1.2}
    \begin{tabular*}{\columnwidth}{@{\extracolsep{\fill}}lcccc@{}}
        \toprule
        & Stage 1 & Stage 2 & Stage 3 & Stage 4 \\
        Resolution
        & $H{\times}W$
        & $\tfrac{H}{2}{\times}\tfrac{W}{2}$
        & $\tfrac{H}{4}{\times}\tfrac{W}{4}$
        & $\tfrac{H}{8}{\times}\tfrac{W}{8}$ \\
        \midrule
        \evlabel{Retained Mass\\(Query-wise, \%)}
        & \evbox{evidenceRed!20}{58.7}
        & \evbox{evidenceGreen!20}{90.7}
        & \evbox{evidenceGreen!20}{85.6}
        & \evbox{evidenceAmber!24}{80.2} \\
        \evlabel{Retained Mass\\($8\times8$ Block-wise, \%)}
        & \evbox{evidenceRed!20}{58.5}
        & \evbox{evidenceGreen!20}{88.0}
        & \evbox{evidenceGreen!20}{79.6}
        & \evbox{evidenceAmber!24}{73.2} \\
        \evlabel{Readout Redundancy\\(\%)}
        & \evbox{evidenceRed!20}{92.87}
        & \evbox{evidenceAmber!24}{64.95}
        & \evbox{evidenceGreen!20}{45.34}
        & \evbox{evidenceGreen!20}{57.67} \\
        \midrule
        \evlabel{Sparse SA\\promising?}
        & {\color{black!60}\evlines{No:\\diffuse}}
        & Yes & Yes
        & {\color{black!60}\evlines{Neutral:\\dense cheap}} \\
        \evlabel{What can be\\simplified?}
        & \evlines{Per-query\\readout} & \evlines{Key\\routing} & \evlines{Key\\routing} & {\color{black!60}\evlines{Keep\\dense}} \\
        \midrule
        \textbf{Design choice}
        & \evlines{\textbf{Fixed}\\\textbf{Kernel}}
        & \multicolumn{2}{c}{\evlines{\textbf{Block-routed}\\\textbf{Sparse SA}}}
        & \textbf{Dense SA} \\
        \bottomrule
    \end{tabular*}
\end{table}

\subsection{Block-Routed Hybrid Attention}
\label{sec:BRHA}

Although our preliminary analyses in Sec.~\ref{sec:sparsity} suggest the potential for sparse attention at intermediate stages, retaining most of the attention mass with sparse attention alone requires a substantial key budget; the evaluated 25\% budget still entails $N^2/4$ query--key interactions for $N$ tokens.
More aggressive sparsification reduces this cost, but restricts fine-grained interactions to fewer spatial regions.
We therefore propose BRHA, which combines dense aggregation of learned summaries from all blocks with sparse aggregation of token-wise residuals within selected blocks, using shared key and value projections.
To exploit spatial redundancy in routing, queries within the same block share the selected key-block indices.

First, LayerNorm~(LN) is applied at each token or stage-wise degradation feature.
A single pointwise projection $W_{qkv}$ of $\operatorname{LN}(\mathbf{X})$ yields token queries, keys, and values $Q,K,V$ shared by the coarse and fine branches.
We partition the resulting $N=H_sW_s$ tokens into $M=N/b^2$ non-overlapping $b\times b$ blocks.
Unless stated otherwise, the equations describe one head with token width $d$; $i,j$ index blocks and $t,u$ index tokens within a block, with $Q_i,K_i,V_i\in\mathbb{R}^{b^2\times d}$.
Separate DWCs $D_q,D_k,D_v$ summarize the query, key, and value blocks, with $V_{\coarseidx}=D_v(V)$.

The context useful for restoration can depend on the degradation, motivating degradation-conditioned block scores.
A $3\times3$ convolution extracts degradation features from the corresponding encoder stage, and a block-wise DWC $D_e$ produces descriptors $E$.
These descriptors are shared across BRHA modules at the same resolution, including the corresponding decoder stage.
All block summarizers use $b\times b$ kernels with stride $b\times b$, initialized with average-pooling kernels and learned jointly with the network.
Each BRHA module forms head-specific embeddings and block queries and keys:
\begin{equation}
    \begin{aligned}
        e &= W_e\!\left(\operatorname{LN}(E)\right),\\
        Q_{\coarseidx} &= \psi_q\!\left([\operatorname{LN}(D_q(Q)),e]\right),\\
        K_{\coarseidx} &= \psi_k\!\left([\operatorname{LN}(D_k(K)),e]\right),
    \end{aligned}
    \label{eq:bsha_fusion}
\end{equation}
where $[\cdot,\cdot]$ denotes channel-wise concatenation and $W_e$ is a head-specific linear projector producing degradation embeddings $e\in\mathbb{R}^{M\times d_r}$.
Separate two-layer MLPs $\psi_q,\psi_k$ with GELU map $d+d_r$ to $d_r$ channels.

\noindent\textbf{Shared scores and coarse aggregation.}
Within each head, a common score matrix $S$ supplies coarse aggregation weights, sparse block selection, and fine-attention gates:
\begin{equation}
    S=\frac{Q_{\coarseidx}K_{\coarseidx}^\top}{\sqrt{d_r}},\qquad O_{\coarseidx}=\operatorname{Softmax}(S)V_{\coarseidx},
    \label{eq:bsha_scores}
\end{equation}
where softmax is applied row-wise over all $M$ blocks.
Let $U$ broadcast each block descriptor to its $b^2$ token positions; $U(O_{\coarseidx})$ then provides a shared global context for the tokens in each query block.

\noindent\textbf{Fine sparse residual aggregation.}
For each query block $i$, we retain its own block and the highest-scoring remote blocks:
\begin{equation}
    \sparseidx=\{i\}\cup\operatorname{TopK}_{j\ne i}(S_{ij},k_r),
    \label{eq:bsha_routing}
\end{equation}
where the remote block budget $k_r$ is capped at $M-1$.
All queries in block $i$ share $\sparseidx$, while retaining individual token-level attention weights.
This block-wise routing sharing improves memory access efficiency~\cite{SVG,SVG2}, significantly accelerating latency.
Next, with residual values $V_r=V-U(V_{\coarseidx})$, fine attention computes:
\begin{equation}
    \begin{aligned}
        A_i&=\operatorname{Softmax}\!\left(
            \frac{Q_iK_{\sparseidx}^{\top}}{\sqrt d}+B_i+L_i
        \right),\\
        [O_{\fineidx}]_i&=A_iV_{r,\sparseidx}.
    \end{aligned}
    \label{eq:bsha_residual}
\end{equation}
Here, $K_{\sparseidx}$ and $V_{r,\sparseidx}$ concatenate the tokens in the selected blocks, and softmax normalizes each query row jointly over those tokens.
$B_i$ repeats the learnable within-block relative position bias $B_{t,u}$ for each selected block, while $L_i$ broadcasts a block-level log gate $\ell_{ij}$ to its token pairs.
Following SAS~\cite{SAS}, we set $\ell_{ii}=0$ for the own block and $\ell_{ij}=\log\operatorname{Softmax}_{j\ne i}(S_{ij})$ for remote blocks, normalized over all remote candidates.
These continuous gates provide a gradient path from restoration to the block scores despite discrete top-$k_r$ selection.
We train BRHA from scratch with these log gates and retain them at inference to preserve the score-conditioned aggregation used during training.

Aggregating raw values $V$ in fine attention would redundantly capture the block-level summary components already represented by the coarse branch.
Instead, our value residualization explicitly subtracts the selected-summary term, yielding $[O_{\fineidx}]_i = A_iV_{\sparseidx} - A_i[U(V_{\coarseidx})]_{\sparseidx}$.
The coarse branch aggregates all-block summaries, while the fine branch uses residual values within selected blocks; their spatial-frequency distributions are compared in Fig.~\ref{fig:brha_value_frequency} of the supplementary material.
Whereas NSA~\cite{NSA} requires separate key--value projections per branch to avoid cross-branch interference, our summary--residual decomposition leverages the spatial continuity of visual features~\cite{liif} to disentangle representations under a single shared projection.
Compared with the separate-K/V variant, BRHA improves PSNR by 0.12\,dB with 24\% fewer attention-module parameters, supporting our summary--residual design~(Table~\ref{tab:attention_ablation}).

\noindent\textbf{Branch fusion.} We combine the two branches for each head and apply a pointwise output projection:

\begin{equation}
    \begin{aligned}
        O^{(h)}&=O_{\fineidx}^{(h)}+\alpha_h U\!\left(O_{\coarseidx}^{(h)}\right),\\
        \mathbf{Y}&=W_p\!\left(\operatorname{Concat}_h(O^{(h)})\right),
    \end{aligned}
    \label{eq:bsha_output}
\end{equation}
where $\alpha_h$ is a learnable head-wise coarse weight and $W_p: C \rightarrow C$ is a pointwise projector.

\noindent\textbf{Complexity and efficient implementation.}
Our BRHA has a complexity of $\mathcal{O}\!\left(N^2(d_r+d)/b^4+N(k_r+1)b^2d\right)$.
At inference, we use $b=8$ and $k_r=16$, requiring only 26.7\% and 6.7\% of the theoretical query--key interactions of a 25\% token-sparse reference for $128\times128$ and $256\times256$ feature maps, respectively.
Despite this reduction, BRHA achieves higher PSNR than Sparse-only with 25\% of KV blocks and shows smaller PSNR drops when $k_r$ is reduced at inference (Figure~\ref{fig:attention_k_sweep} in the supplementary materials).

Still, theoretical computational complexity alone does not fully capture latency, which also depends on memory access and GPU utilization~\cite{FA2, FasterNet, SHViT, ESC, PLKSR}.
We therefore implement BRHA in Triton~\cite{Triton} using online softmax~\cite{FA2} to avoid storing the block-score matrix $S$.
Routing merges tile-local top-$k_r$ candidates, and fine attention gathers only the routed blocks $\sparseidx$.
Value residuals are computed on the fly, avoiding materialization of $V_r$.

\begin{table*}[t]
\centering
\scriptsize
\fboxsep0.75pt
\setlength\tabcolsep{1.55pt}
\caption{Comparison on CDD-11 in PSNR (dB, $\uparrow$) and \colorbox{clblue!50}{SSIM ($\uparrow$)}.
\textbf{Bold} marks the best results.}
\vspace{-0.2cm}
\label{tab:cdd11:higlr}
\resizebox{\textwidth}{!}{\begin{tabular}{@{}l@{\hspace{6pt}}c@{\hspace{6pt}}*{12}{r@{\hspace{1pt}}>{\columncolor{clblue!50}[0.5pt][\tabcolsep]}l}@{}}
\toprule
\multirow{2}{*}{Method}
& \multirow{2}{*}{Params.}
& \multicolumn{8}{c}{\textit{CDD11-Single}}
& \multicolumn{10}{c}{\textit{CDD11-Double}}
& \multicolumn{4}{c}{\textit{CDD11-Triple}}
& \multicolumn{2}{c}{\multirow{2}{*}{Avg.}}\\
\cmidrule(lr){3-10}
\cmidrule(lr){11-20}
\cmidrule(lr){21-24}
&& \multicolumn{2}{c}{Low~(L)}
& \multicolumn{2}{c}{Haze~(H)}
& \multicolumn{2}{c}{Rain~(R)}
& \multicolumn{2}{c}{Snow~(S)}
& \multicolumn{2}{c}{L+H}
& \multicolumn{2}{c}{L+R}
& \multicolumn{2}{c}{L+S}
& \multicolumn{2}{c}{H+R}
& \multicolumn{2}{c}{H+S}
& \multicolumn{2}{c}{L+H+R}
& \multicolumn{2}{c}{L+H+S}\\
\midrule

AirNet~\cite{AirNet}
& 9M
& 24.83 & \cc{.778}
& 24.21 & \cc{.951}
& 26.55 & \cc{.891}
& 26.79 & \cc{.919}
& 23.23 & \cc{.779}
& 22.82 & \cc{.710}
& 23.29 & \cc{.723}
& 22.21 & \cc{.868}
& 23.29 & \cc{.901}
& 21.80 & \cc{.708}
& 22.24 & \cc{.725}
& 23.75 & \cc{.814}\\

WGWSNet~\cite{WGWSNet}
& 26M
& 24.39 & \cc{.774}
& 27.90 & \cc{.982}
& 33.15 & \cc{.964}
& 34.43 & \cc{.973}
& 24.27 & \cc{.800}
& 25.06 & \cc{.772}
& 24.60 & \cc{.765}
& 27.23 & \cc{.955}
& 27.65 & \cc{.960}
& 23.90 & \cc{.772}
& 23.97 & \cc{.771}
& 26.96 & \cc{.863}\\

WeatherDiff~\cite{WeatherDiff}
& 83M
& 23.58 & \cc{.763}
& 21.99 & \cc{.904}
& 24.85 & \cc{.885}
& 24.80 & \cc{.888}
& 21.83 & \cc{.756}
& 22.69 & \cc{.730}
& 22.12 & \cc{.707}
& 21.25 & \cc{.868}
& 21.99 & \cc{.868}
& 21.23 & \cc{.716}
& 21.04 & \cc{.698}
& 22.49 & \cc{.799}\\

PromptIR~\cite{PromptIR}
& 36M
& 26.32 & \cc{.805}
& 26.10 & \cc{.969}
& 31.56 & \cc{.946}
& 31.53 & \cc{.960}
& 24.49 & \cc{.789}
& 25.05 & \cc{.771}
& 24.51 & \cc{.761}
& 24.54 & \cc{.924}
& 23.70 & \cc{.925}
& 23.74 & \cc{.752}
& 23.33 & \cc{.747}
& 25.90 & \cc{.850}\\

OneRestore~\cite{Onerestore}
& 6M
& 26.48 & \cc{.826}
& 32.52 & \cc{.990}
& 33.40 & \cc{.964}
& 34.31 & \cc{.973}
& 25.79 & \cc{.822}
& 25.58 & \cc{.799}
& 25.19 & \cc{.789}
& 29.99 & \cc{.957}
& 30.21 & \cc{.964}
& 24.78 & \cc{.788}
& 24.90 & \cc{.791}
& 28.47 & \cc{.878}\\

MoCE-IR-S~\cite{MOCEIR}
& 11M
& 27.26 & \cc{.824}
& 32.66 & \cc{.990}
& 34.31 & \cc{.970}
& 35.91 & \cc{.980}
& 26.24 & \cc{.817}
& 26.25 & \cc{.800}
& 26.04 & \cc{.793}
& 29.93 & \cc{.964}
& 30.19 & \cc{.970}
& 25.41 & \cc{.789}
& 25.39 & \cc{.790}
& 29.05 & \cc{.881}\\

BioIR-T~\cite{BioIR}
& 1M
& 27.14 & \cc{.834}
& 34.77 & \cc{.992}
& 34.48 & \cc{.970}
& 36.26 & \cc{.979}
& 26.11 & \cc{.830}
& 26.17 & \cc{.809}
& 26.01 & \cc{.804}
& 31.64 & \cc{.968}
& 31.60 & \cc{.972}
& 25.22 & \cc{.799}
& 25.34 & \cc{.802}
& 29.52 & \cc{.887}\\

BioIR-B~\cite{BioIR}
& 7M
& 27.52 & \cc{.838}
& 36.95 & \cc{.995}
& 35.35 & \cc{.974}
& 37.28 & \cc{.981}
& 26.79 & \cc{.835}
& 26.78 & \cc{.817}
& 26.63 & \cc{.812}
& 32.77 & \cc{.973}
& 33.39 & \cc{.977}
& 26.20 & \cc{.811}
& 26.28 & \cc{.811}
& 30.54 & \cc{.893}\\

AnyIR-S~\cite{AnyIR}
& 10M
& 27.47 & \cc{.835}
& 34.38 & \cc{.992}
& 34.64 & \cc{.971}
& 36.21 & \cc{.981}
& 26.54 & \cc{.830}
& 26.49 & \cc{.812}
& 26.45 & \cc{.805}
& 30.98 & \cc{.967}
& 31.55 & \cc{.972}
& 25.82 & \cc{.804}
& 26.01 & \cc{.805}
& 29.69 & \cc{.889}\\

MIRAGE-T~\cite{MIRAGE}
& 6M
& 27.13 & \cc{.830}
& 32.39 & \cc{.989}
& 34.23 & \cc{.969}
& 35.57 & \cc{.978}
& 26.04 & \cc{.823}
& 26.21 & \cc{.807}
& 26.07 & \cc{.799}
& 29.49 & \cc{.962}
& 29.72 & \cc{.967}
& 25.17 & \cc{.793}
& 25.41 & \cc{.793}
& 28.86 & \cc{.883}\\

MIRAGE-S~\cite{MIRAGE}
& 10M
& 27.41 & \cc{.833}
& 33.12 & \cc{.992}
& 34.66 & \cc{.971}
& 35.98 & \cc{.981}
& 26.55 & \cc{.828}
& 26.53 & \cc{.810}
& 26.33 & \cc{.803}
& 30.32 & \cc{.965}
& 30.27 & \cc{.969}
& 25.59 & \cc{.801}
& 25.86 & \cc{.799}
& 29.33 & \cc{.887}\\

R2R~\cite{R2R}
& 20M
& 27.29 & \cc{.836}
& 32.88 & \cc{.992}
& 35.12 & \cc{.971}
& 36.54 & \cc{.981}
& 26.21 & \cc{.833}
& 26.70 & \cc{.816}
& 26.33 & \cc{.809}
& 30.41 & \cc{.966}
& 30.37 & \cc{.971}
& 25.47 & \cc{.804}
& 25.21 & \cc{.803}
& 29.32 & \cc{.889}\\

CEA-IR-S~\cite{CEAIR}
& 10M
& 27.47 & \cc{.827}
& 34.75 & \cc{.993}
& 34.72 & \cc{.971}
& 36.49 & \cc{.982}
& 26.67 & \cc{.824}
& 26.65 & \cc{.806}
& 26.58 & \cc{.802}
& 31.15 & \cc{.968}
& 31.49 & \cc{.974}
& 25.87 & \cc{.797}
& 25.93 & \cc{.799}
& 29.79 & \cc{.886}\\

\midrule
HiGLR (\textit{ours})
& 7M
& \textbf{27.80} & \cc{\textbf{.841}}
& \textbf{40.73} & \cc{\textbf{.997}}
& \textbf{36.04} & \cc{\textbf{.978}}
& \textbf{37.81} & \cc{\textbf{.983}}
& \textbf{27.50} & \cc{\textbf{.840}}
& \textbf{27.10} & \cc{\textbf{.822}}
& \textbf{27.05} & \cc{\textbf{.816}}
& \textbf{34.38} & \cc{\textbf{.977}}
& \textbf{35.23} & \cc{\textbf{.980}}
& \textbf{26.73} & \cc{\textbf{.816}}
& \textbf{26.83} & \cc{\textbf{.817}}
& \textbf{31.56} & \cc{\textbf{.897}}\\
\bottomrule
\end{tabular}}
\vspace{-0.3cm}
\end{table*}

\begin{figure}[t]
    \centering
    \includegraphics[width=\linewidth]{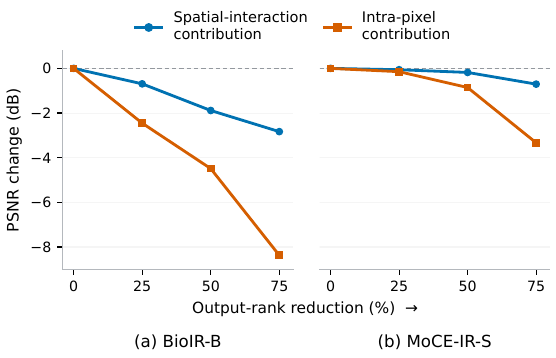}
    \vspace{-0.5cm}
    \caption{Sensitivity to output-rank reduction of spatial-interaction and intra-pixel contributions in pretrained BioIR-B and MoCE-IR-S~\cite{BioIR,MOCEIR}, evaluated on CDD-11.
    Each contribution is independently compressed using PCA~\cite{zhang_cvpr2015_response_pca}.
    Curves show PSNR changes relative to the uncompressed models.
    Please refer to Sec.~\ref{sup_sec:compression_details} for experimental details.}
    \label{fig:ffn-width}
    \vspace{-0.4cm}
\end{figure}

\subsection{Dual-Space Feed-Forward Network}
\label{sec:DSFFN}
\noindent{\textbf{Spatial-Mixing Redundancy in ConvFFNs.}}
At HR stages, eliminating substantial activation memory traffic requires kernel fusion to avoid materializing expanded FFN features~(Sec.~\ref{sub_sec:ffn_memory_bound}).
However, ConvFFNs perform DWC directly in the expanded feature space, introducing cross-pixel dependencies that complicate effective fusion.
To mitigate the overhead of intermediate spatial mixing, we investigate whether this expanded width is strictly necessary, examining whether spatial interactions tolerate greater dimensionality reduction than intra-pixel contributions.

We independently compress either output contribution using PCA~\cite{zhang_cvpr2015_response_pca} in pretrained BioIR-B~\cite{BioIR} and MoCE-IR-S~\cite{MOCEIR} on CDD-11~\cite{Onerestore}, keeping the complementary contribution, hidden widths, and model weights unchanged.
More details are provided in Sec.~\ref{sup_sec:compression_details}.
As shown in Fig.~\ref{fig:ffn-width}, spatial-interaction compression incurs smaller PSNR losses in these two models, with the gap widening under stronger compression.
At 50\% output-rank reduction, BioIR-B and MoCE-IR-S lose 1.89 and 0.18\,dB, respectively, compared with 4.49 and 0.86\,dB for intra-pixel compression.

\vspace{0.2cm}

\noindent{\textbf{DSFFN design.}} These output-space observations motivate decoupling spatial interaction from nonlinear feature expansion: computing spatial interaction in a compact space and lifting its result into the expanded feature space, rather than requiring both operations to share the same width.

Motivated by this observation, DSFFN combines a compact spatial path with an expanded pointwise path.
Given $\mathbf{X}$, the two paths and their output are
\begin{equation}
    \setlength{\arraycolsep}{2pt}
    \renewcommand{\arraystretch}{1.25}
    \begin{array}{>{\displaystyle}r>{\displaystyle}l}
        T &{}= D\!\left(\phi\!\left(\operatorname{RMSNorm}\!\left(W_r(\mathbf{X})\right)\right)\right),\\
        \rowcolor{dsffnorange!45}
        {[Z,G]} &{}= \operatorname{Split}\!\left(W_{zg}\!\left(\operatorname{LN}(\mathbf{X})\right)\right),\\
        \rowcolor{dsffnorange!45}
        \mathbf{Y} &{}= W_o\!\left(\phi\!\left(Z+W_s(T)\right)\odot\sigma(G)\right),
    \end{array}
    \label{eq:dsffn_output}
\end{equation}
where the pointwise projectors are $W_r:C\to R$ with $R<\gamma C$, $W_{zg}:C\to2\gamma C$, $W_s:R\to\gamma C$, and $W_o:\gamma C\to C$; $\gamma$ is the per-branch channel expansion ratio.
$D$ is a $3\times3$ DWC on $R$ channels, and $Z,G\in\mathbb{R}^{H\times W\times\gamma C}$ are obtained by an equal channel-wise split.
Here $\phi$ is GELU, $\sigma$ is sigmoid, and $\odot$ denotes element-wise multiplication.
The spatial path takes $\mathbf{X}$ before LN; following subspace normalization methods~\cite{DeepSeekV2, NoRA}, we apply RMSNorm to regulate the compressed features' scale before GELU and spatial mixing.
We empirically choose RMSNorm over LayerNorm to avoid per-token mean subtraction that may remove cues useful for spatial mixing by DWC.

Unlike ConvFFN, which applies DWC to both expanded branches, DSFFN injects compact spatial features into the value path while keeping the gate pointwise, decoupling the spatial width $R$ from both the input width $C$ and the expanded width $\gamma C$.
A comparison with a sequential design that applies DWC before pointwise expansion~\cite{ConvNeXt} is provided in Section~\ref{supp:ffn-traffic} of the supplementary material.

\begin{table}[!t]
\centering
\footnotesize
\fboxsep0.5pt
\setlength\tabcolsep{3pt}
\caption{Average PSNR (dB, $\uparrow$) / \colorbox{clblue!50}{SSIM ($\uparrow$)} on AIO-3 and AIO-5. Full per-task results are in Tables~\ref{tab:aio3:higlr} and~\ref{tab:aio5:higlr} in the supplementary material. \textbf{Bold} marks the best results (ties included). $^{\dagger}$Additional frozen CLIP encoder: 88M parameters.}
\vspace{-0.2cm}
\label{tab:aio_summary:higlr}
\begin{tabular}{@{}lcc>{\columncolor{clblue!50}}cc>{\columncolor{clblue!50}}c@{}}
\toprule
\multirow{2}{*}{Method} & \multirow{2}{*}{Params.} & \multicolumn{2}{c}{AIO-3} & \multicolumn{2}{c}{AIO-5} \\
\cmidrule(lr){3-4}\cmidrule(lr){5-6}
& & PSNR & SSIM & PSNR & SSIM \\
\midrule
AirNet~\cite{AirNet} & 9M & 31.20 & .910 & 25.49 & .847 \\
PromptIR~\cite{PromptIR} & 36M & 32.06 & .913 & 29.15 & .904 \\
VLU-Net~\cite{VLUNet} & 35M$^{\dagger}$ & 32.70 & .919 & 30.11 & .905 \\
MoCE-IR-S~\cite{MOCEIR} & 11M & 32.57 & .916 & 30.08 & .913 \\
MoCE-IR~\cite{MOCEIR} & 25M & 32.73 & .917 & 30.58 & .919 \\
BioIR-L~\cite{BioIR} & 16M & 32.87 & .921 & 30.99 & .922 \\
AnyIR-S~\cite{AnyIR} & 9M & 32.86 & .920 & 30.75 & .916 \\

MIRAGE-T~\cite{MIRAGE} & 6M & 32.77 & .919 & 30.41 & .912 \\
MIRAGE-S~\cite{MIRAGE} & 10M & 32.91 & .919 & 30.68 & .914 \\
R2R~\cite{R2R} & 20M & 32.53 & .918 & 30.48 & .921 \\

CEA-IR-S~\cite{CEAIR} & 10M & 32.89 & .918 & 30.80 & .919 \\
CEA-IR~\cite{CEAIR} & 22M & 33.07 & \textbf{.922} & 31.13 & .922 \\
\midrule
HiGLR (\textit{ours}) & 7M & \textbf{33.14} & .921 & \textbf{31.30} & \textbf{.926} \\
\bottomrule
\end{tabular}
\vspace{-0.3cm}
\end{table}

\noindent\textbf{Efficient implementation.}
A naive ConvFFN materializes expanded features in global memory, while tiled fusion must handle spatial halos and may recompute expanded features in overlapping regions.
DSFFN confines these spatial dependencies to $R$ channels, leaving the expanded-space operations pointwise.
We fuse the spatial up-projection $W_s$ with the pointwise path, activation, gating, and output projection (highlighted in {\setlength{\fboxsep}{1pt}\colorbox{dsffnorange!45}{light orange}} in Eq.~\eqref{eq:dsffn_output}), avoiding expanded intermediate materialization while preserving nonlinear transformations.
Details are provided in Sec.~\ref{supp:dsffn_impl}.

\section{Experiments}
In this section, we evaluate HiGLR's restoration quality, efficiency, and key design choices.
\textit{Implementation and training details are provided in Sections~\ref{sup_sec:implementation_details} and~\ref{sup_sec:training_details}.}

\subsection{Quantitative Results}
To measure restoration across task mixtures and compound degradations, we compare HiGLR with representative AIO-IR methods~\cite{MOCEIR,BioIR,MIRAGE,R2R,CEAIR} on AIO-3~\cite{AirNet}, AIO-5~\cite{R2R}, and CDD-11~\cite{Onerestore}, reporting PSNR and SSIM measured in the RGB color space following previous work~\cite{R2R}.
On AIO-3 and AIO-5, HiGLR averages 33.14 and 31.30\,dB, exceeding CEA-IR by 0.07 and 0.17\,dB with 7M versus 22M parameters~(Table~\ref{tab:aio_summary:higlr}).
On CDD-11, HiGLR averages 31.56\,dB, 1.02\,dB above BioIR-B, with the highest PSNR and SSIM in all eleven conditions~(Table~\ref{tab:cdd11:higlr}).
These results demonstrate the superior performance of HiGLR and highlight the importance of processing HR features.

\begin{table}[t]
  \centering
  \caption{CDD-11 average PSNR and inference costs ($512\times512$, batch 1, RTX~4090). Peak: allocated GPU memory; traffic: DRAM reads plus writes measured with NCU.}
  \label{tab:efficiency}
  \vspace{-0.2cm}
  \footnotesize
  \setlength{\tabcolsep}{2.5pt}
  \renewcommand{\arraystretch}{1.08}
  \begin{tabular}{@{}lrrrrr@{}}
    \toprule
    Method & Params & Latency & Peak & Traffic & PSNR \\
           & (M)$\downarrow$ & (ms)$\downarrow$ & (MiB)$\downarrow$ & (GB)$\downarrow$ & (dB)$\uparrow$ \\
    \midrule
    MoCE-IR-S~\cite{MOCEIR} & 11.48 & 76.44 & 661.1 & 39.98 & 29.05 \\
    BioIR-B~\cite{BioIR} & 7.28 & 89.01 & 1401.9 & 68.96 & 30.54 \\
    MIRAGE-S~\cite{MIRAGE} & 9.68 & 83.48 & 1254.6 & 48.62 & 29.33 \\
    R2R~\cite{R2R} & 19.77 & 31.76 & 422.0 & 4.92 & 29.32 \\
    CEA-IR-S~\cite{CEAIR} & 9.62 & 96.07 & 861.9 & 51.28 & 29.79 \\
    \midrule
    \rowcolor{tablehighlight}
    HiGLR~(Ours) & \textbf{7.19} & \textbf{22.64} & \textbf{294.9} & \textbf{4.82} & \textbf{31.56} \\
    \bottomrule
  \end{tabular}
\end{table}

\subsection{Efficiency Comparison}
To measure inference speed and memory efficiency, we compare latency, peak GPU memory, and DRAM traffic against representative AIO-IR methods~\cite{MOCEIR, BioIR, MIRAGE, R2R, CEAIR}.
Following prior methods, we use FP32 precision by default, with BF16 selectively employed only for attention core computations~\cite{FA2}.
We measure DRAM traffic as total reads and writes using NVIDIA Nsight Compute (NCU).
At $512\times512$ on an RTX~4090, HiGLR requires only 22.64\,ms and 294.9\,MiB of peak allocated memory, running $3.93\times$ faster with 79.0\% lower peak memory than BioIR-B~(Table~\ref{tab:efficiency}).
Despite using 20 DSFFN blocks at finest-scale stages, HiGLR achieves the lowest latency and DRAM traffic (4.82\,GB) among the compared methods, highlighting the efficiency of our DSFFN.

\subsection{Ablation Study}
We conduct ablation studies to evaluate the contributions of HiGLR's key design choices to restoration performance and efficiency.
\textit{Unless otherwise specified, all ablation models are trained on CDD-11 for 120 epochs, half the duration of the full training schedule. Consequently, the results may differ from those reported in Table~\ref{tab:cdd11:higlr}.}

\begin{table}[t]
\centering
\caption{Global modeling operators: CDD-11 PSNR with each operator in \protect\colorbox{tablehighlight}{Stages~2 and~3,} and per-stage single attention module latency~($512{\times}512$, batch 1, RTX 4090, FP32 projection and BF16 for attention core).
Ind.: query-wise routing.}
\label{tab:stage_attention_latency}
\label{tab:attention_comparison}
\vspace{-0.2cm}
\setlength{\tabcolsep}{2.5pt}
\renewcommand{\arraystretch}{1.15}
\footnotesize

\begin{tabular}{@{}lc>{\columncolor{tablehighlight}}c>{\columncolor{tablehighlight}}ccc@{}}
\toprule
 & \multicolumn{4}{c}{Latency (ms)$\downarrow$} & PSNR$\uparrow$ \\
\cmidrule(lr){2-5}
Method & Stage 1 & Stage 2 & Stage 3 & Stage 4 & CDD-11 \\
\midrule
Resolution
 & $H{\times}W$
 & $\frac{H}{2}{\times}\frac{W}{2}$
 & $\frac{H}{4}{\times}\frac{W}{4}$
 & $\frac{H}{8}{\times}\frac{W}{8}$ & \\
$C/h/d$
 & 32/1/32 & 64/1/64 & 128/2/64 & 256/4/64 & \\
\midrule
Dense SA~\cite{Attention}
 & 59.859 & 7.080 & 1.481 & \underline{0.464} & 30.20 \\
MDTA~\cite{Restormer}
 & \textbf{0.893} & \textbf{0.495} & \textbf{0.483} & 0.476 & 29.04 \\
RELA~\cite{RELA}
 & \underline{0.930} & \underline{0.525} & \underline{0.496} & \textbf{0.458} & 30.12 \\
BRHA-Ind.
 & 29.450 & 7.485 & 3.907 & 2.452 & \textbf{30.39} \\
BRHA (ours)
 & 2.206 & 1.425 & 1.163 & 1.030 & \underline{30.37} \\
\bottomrule
\end{tabular}
\vspace{-0.2cm}
\end{table}

\paragraph{Global modeling and routing.}
To evaluate global modeling quality and routing efficiency, we compare global operators and routing granularity in Stages~2 and~3~(Table~\ref{tab:attention_comparison}).
Compared with dense self-attention, BRHA reaches 30.37 rather than 30.20\,dB and reduces Stage~2 latency from 7.080 to 1.425\,ms.
BRHA also achieves higher PSNR than the evaluated sub-quadratic alternatives, MDTA~\cite{Restormer} and RELA~\cite{RELA}, with little increase in latency.
Compared with query-wise routing, block-wise sharing changes PSNR from 30.39 to 30.37\,dB while reducing Stage~2 and Stage~3 latency by $5.25\times$ and $3.36\times$, respectively, suggesting that neighboring queries can share key-block selections.

\begin{table}[t]
\centering
\caption{Attention component ablation on CDD-11.
$V_r$ and $l$ denote value residualization and block-level log gates, respectively.}
\label{tab:attention_ablation}
\vspace{-0.2cm}
\footnotesize
\begin{tabular}{cccccc}
\toprule
Coarse & Sparse & Separate $K$/$V$ & $V_r$ & log gate & PSNR$\uparrow$ \\
\midrule
-- & $\checkmark$ & -- & -- & $\checkmark$ & 29.98 \\
$\checkmark$ & $\checkmark$ & -- & -- & $\checkmark$ & 29.95 \\
$\checkmark$ & -- & -- & -- & -- & 30.02 \\
$\checkmark$ & $\checkmark$ & -- & $\checkmark$ & -- & 30.18 \\
$\checkmark$ & $\checkmark$ & $\checkmark$ & -- & $\checkmark$ & 30.25 \\
\rowcolor{tablehighlight}
$\checkmark$ & $\checkmark$ & -- & $\checkmark$ & $\checkmark$ & 30.37 \\
\bottomrule
\end{tabular}
\end{table}

\paragraph{BRHA design ablation.}
To evaluate the effectiveness of the proposed BRHA components, we ablate its branches, value residualization, and log gates~(Table~\ref{tab:attention_ablation}).
BRHA achieves 30.37\,dB, outperforming sparse-only and coarse-only attention by 0.39 and 0.35\,dB, respectively.
Notably, residualizing fine-scale values improves PSNR by 0.42\,dB over simply combining coarse and sparse attention on raw values, and by 0.12\,dB over using separate $K$/$V$ projections for the two branches.
These results support explicitly separating coarse summaries from fine residuals rather than relying on branch combination or separate projections alone to learn complementary representations.
Removing log gates further reduces PSNR to 30.18\,dB, supporting the use of routing scores to modulate fine-scale aggregation.

\begin{figure}[t]
\centering
{\setlength{\fboxsep}{3pt}\includegraphics[width=\dimexpr\linewidth-6pt\relax]{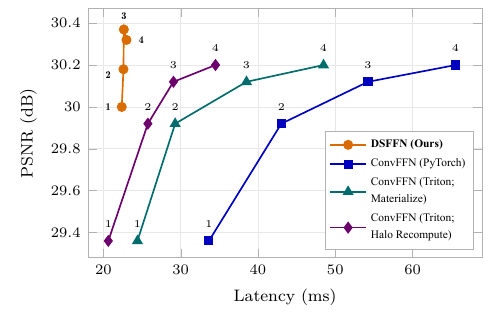}}
\vspace{-0.4cm}
\caption{FFN ablation studies on CDD-11 ($512\times512$, batch 1, RTX~4090).
Numbers denote per-branch expansion $\boldsymbol{\gamma}$.
The full comparison is in supplementary Table~\ref{tab:supp-ffn-full}.}
\vspace{-0.2cm}
\label{fig:ffn-latency-psnr}
\end{figure}

\begin{figure}[t]
    \centering
    \includegraphics[width=\linewidth]{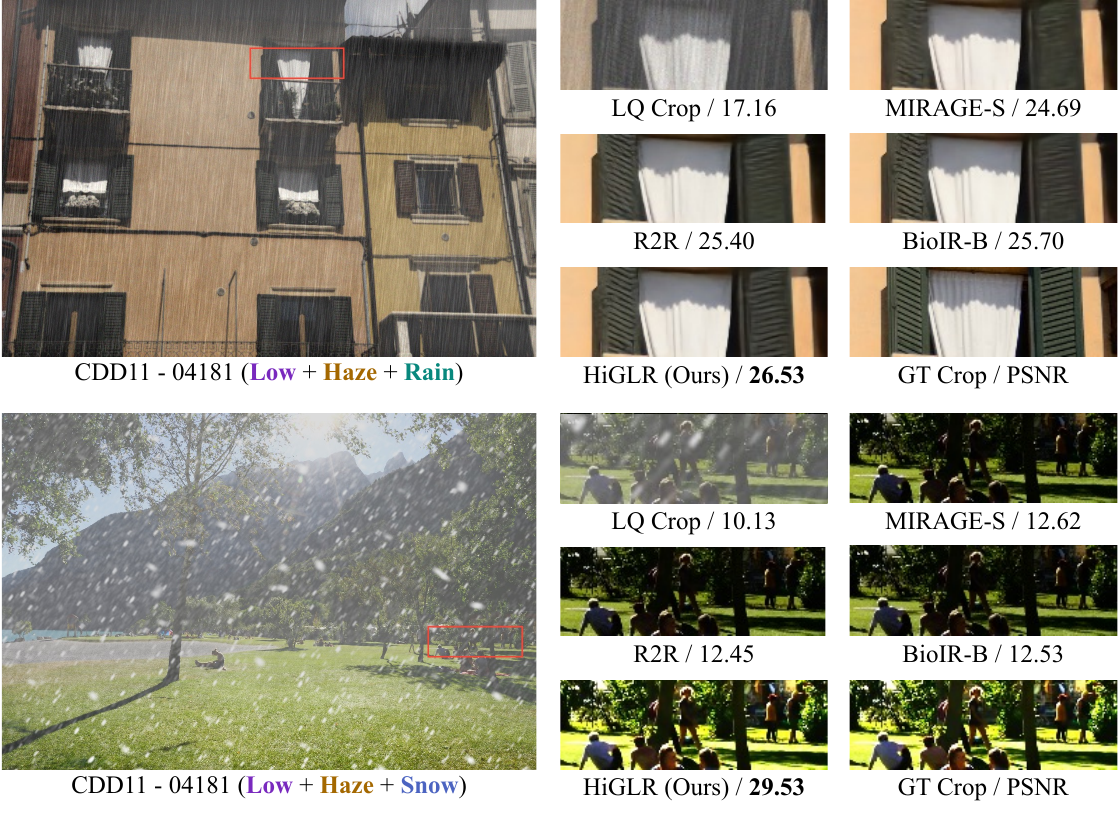}
    \caption{
    Visual comparison on CDD-11 under compound degradations. Red boxes indicate the enlarged regions. Please refer to Section~\ref{sup_sec:more_visual_results} for more visual results.
    }
    \vspace{-0.2cm}
    \label{fig:main_visual}
\end{figure}

\begin{table}[t]
  \centering
  \caption{DSFFN spatial-rank ablation on CDD-11 with $\gamma=3$.
  }
  \label{tab:dsffn_rank}
  \vspace{-0.2cm}
  \renewcommand{\arraystretch}{1.0}
  \footnotesize
  \setlength{\tabcolsep}{3.5pt}
  \begin{tabular}{@{}rrrrrr@{}}
    \toprule
    Rank $R$ & Params & Latency & Peak & Traffic & PSNR \\
             & (M) & (ms) & (MiB) & (GB) & (dB) \\
    \midrule
    8 & 7.17 & 22.69 & 294.9 & 4.76 & 29.95 \\
    \rowcolor{tablehighlight}
    16 & 7.19 & 22.64 & 294.9 & 4.82 & 30.37 \\
    32 & 7.24 & 23.72 & 295.1 & 4.73 & 30.29 \\
    48 & 7.28 & 25.96 & 295.3 & 4.77 & 30.33 \\
    64 & 7.33 & 25.55 & 295.4 & 4.73 & 30.39 \\
    \bottomrule
  \end{tabular}
\end{table}

\paragraph{FFN expansion ratio and comparison with ConvFFN.}
We compare DSFFN with ConvFFN across expansion ratios to measure how $\gamma$ affects the quality--latency trade-off.
As $\gamma$ increases from 1 to 4, DSFFN's full-model latency stays within 22.37--22.98\,ms, whereas halo-fused ConvFFN latency rises from 20.63 to 34.50\,ms~(Fig.~\ref{fig:ffn-latency-psnr}).
At $\gamma=3$, DSFFN achieves 30.37\,dB at 22.64\,ms versus 30.12\,dB at 29.06\,ms for halo-fused ConvFFN.
These results suggest that decoupling spatial mixing from feature expansion allows wider pointwise transformations with limited growth in inference latency by leveraging kernel fusion.

\paragraph{Spatial rank.}
We vary spatial rank $R$ only in Stage~1 at $\gamma=3$ to measure DSFFN's spatial capacity~(Table~\ref{tab:dsffn_rank}).
The ranks of the remaining stages are fixed, since scaling $R$ proportionally in deeper stages would substantially increase the parameter count.
Increasing $R$ from 8 to 16 raises PSNR from 29.95 to 30.37\,dB at nearly unchanged latency; larger ranks reach 30.29--30.39\,dB with higher latency.
Therefore, we adopt $R=16$ to balance quality and latency.

\begin{table}[t]
\centering
\caption{Effect of training and test budgets $k_r$ on CDD-11. Latency denotes total attention-module latency.}
\label{tab:train_test_k}
\vspace{-0.2cm}
\setlength{\tabcolsep}{4pt}
\renewcommand{\arraystretch}{1.0}
\footnotesize
\begin{tabular}{@{}lcccc@{}}
\toprule
 & \multicolumn{4}{c}{Test $k_r$} \\
\cmidrule(lr){2-5}
Train $k_r$ & 8 & 16 & 32 & 48 \\
\midrule
8 & 30.18 & 30.19 & 30.19 & 30.18 \\
16 & 30.12 & 30.24 & 30.25 & 30.25 \\
32 & 30.34 & \cellcolor{tablehighlight}\underline{30.37} & \underline{30.37} & \textbf{30.40} \\
48 & 30.21 & 30.29 & 30.34 & 30.34 \\
\midrule
Latency (ms) & \textbf{10.11} & \underline{11.01} & 12.35 & 15.32 \\
\bottomrule
\end{tabular}
\vspace{-0.4cm}
\end{table}

\vspace{0.2cm}
\noindent\textbf{Training and inference budgets.}
We vary the remote-block budget $k_r$ independently during training and inference to examine its effect on restoration quality and attention cost~(Table~\ref{tab:train_test_k}).
Among the tested training budgets, $k_r=32$ yields the highest PSNR at every inference budget; increasing the training budget to 48 does not improve performance.
For the model trained with $k_r=32$, increasing inference $k_r$ from 8 to 16 raises PSNR from 30.34 to 30.37\,dB while attention latency increases from 10.11 to 11.01\,ms.
Further increasing inference $k_r$ to 32 or 48 raises attention latency to 12.35 or 15.32\,ms, with little further improvement in PSNR.
We therefore adopt training $k_r=32$ and inference $k_r=16$ to balance quality and cost.

\subsection{Visual Results}
To examine whether the quantitative gains are reflected in visual quality, we compare restored images under compound degradations.
As shown in Figure~\ref{fig:main_visual}, HiGLR restores fine details and colors more accurately than the compared methods.
These results demonstrate that HiGLR also reconstructs images with superior visual quality.

\section{Conclusion}
\vspace{-0.2cm}
In this paper, we presented HiGLR, an efficient multi-scale network for AIO-IR tasks.
Our analysis identified redundant attention readouts at the finest-scale stage and key routing across spatial queries.
We also observed greater tolerance to rank reduction in spatial mixing than in pointwise nonlinear transformation.
These observations motivated local mixing at the finest scale, BRHA with shared routing and complementary coarse and residual attention, and DSFFN with compact spatial mixing and expanded pointwise transformations.
Together, these designs achieve state-of-the-art performance on AIO-3, AIO-5, and CDD-11 while significantly reducing inference latency and memory usage.

\clearpage
\maketitlesupplementary

\renewcommand{\theHsection}{supp.\arabic{section}}

\addtocontents{toc}{\protect\setcounter{tocdepth}{2}}
\begingroup
\makeatletter
\patchcmd{\l@section}{1.5em}{2.4em}{}{}
\patchcmd{\l@subsection}{2.3em}{3.2em}{}{}
\renewcommand{\numberline}[1]{\makebox[\@tempdima][r]{#1\hspace{1em}}}
\tableofcontents
\endgroup
\medskip

\section{Memory-Bound Behavior of Narrow FFNs}
\label{sub_sec:ffn_memory_bound}

\paragraph{Arithmetic intensity.}
Although FFNs consist primarily of matrix multiplications, they are not necessarily compute-bound.
In image restoration (IR), finest-resolution features have many spatial tokens but relatively few channels, making activation movement potentially significant.

Consider an FFN with expansion ratio two, $Y=\phi(XW_1)W_2$, where $X\in\mathbb{R}^{N\times d}$.
The two projections require $8Nd^2$ FLOPs.
Assuming FP32 storage, materialized intermediate activations, and one read per weight, the arithmetic intensity is
\begin{equation}
    \mathcal{I}_{\mathrm{unfused}}
    = \frac{8Nd^2}{4(10Nd+4d^2)}
    = \frac{Nd}{5N+2d}
    \xrightarrow{N\gg d} \frac{d}{5}.
\end{equation}
For $d=32$ and $64$, this yields only $6.4$ and $12.8$ FLOP/byte, respectively.
Thus, increasing spatial resolution does not necessarily improve arithmetic intensity, and narrow FFNs can become sensitive to memory movement.
Even TF32 increases available compute throughput without reducing FP32 tensor storage.

\paragraph{Fusion experiment.}
We benchmark unfused PyTorch and fused Triton implementations of a Linear--GELU--Linear FFN~(w/ expansion ratio two) on an RTX 4090, with batch size one and $512\times512$ spatial resolution.
Both use FP32 storage with TF32 enabled: PyTorch permits TF32, and the fused Triton kernel explicitly uses TF32 dot products.
We report median latency over 300 CUDA-event samples~(three rounds of 100) under CUDA Graph replay, excluding compilation, weight packing, warm-up, and profiling.

\begin{table}[t]
\centering
\small
\setlength{\tabcolsep}{3pt}
\caption{Plain FFN fusion performance (expansion$=2$, $512{\times}512$, batch 1, RTX 4090; FP32/TF32).
U/F: unfused/fused; peak: incremental allocated memory.}
\label{tab:plain_mlp_fusion_4090}
\begin{tabular}{@{}rrrrrr@{}}
\toprule
$d$ & GMACs & \shortstack{U\\(ms)} & \shortstack{F\\(ms)} & Speedup & \shortstack{Peak U/F\\(MiB)} \\
\midrule
32 & 1.074 & 0.333 & 0.066 & $5.05\times$ & 128 / 32 \\
64 & 4.295 & 0.750 & 0.166 & $4.53\times$ & 256 / 64 \\
\bottomrule
\end{tabular}
\end{table}

\begin{table*}[t]
  \centering
  \caption{Stage-wise attention statistics across encoder depths, measured on the same 75 crops as Table~\ref{tab:attention-mass-design}.
  All values are percentages. Retained mass uses a 25\% key budget; block-wise routing shares keys within $8\times8$ query blocks.
  Readout redundancy uses the equal-layer aggregation defined in Sec.~\ref{sup_sec:stagewise_statistics}.
  A dash denotes a stage absent from the encoder.}
  \label{tab:stagewise_depth}
  \vspace{-0.2cm}
  \small
  \setlength{\tabcolsep}{4pt}
  \renewcommand{\arraystretch}{1.15}
  \begin{tabular*}{0.72\textwidth}{@{\extracolsep{\fill}}llccccc@{}}
    \toprule
    \multirow{2}{*}{Encoder depth} & \multirow{2}{*}{Statistic (\%)}
      & Stage 1 & Stage 2 & Stage 3 & Stage 4 & Stage 5 \\
    & & $H\times W$ & $\frac{H}{2}\times\frac{W}{2}$
      & $\frac{H}{4}\times\frac{W}{4}$ & $\frac{H}{8}\times\frac{W}{8}$
      & $\frac{H}{16}\times\frac{W}{16}$ \\
    \midrule
    \multirow{3}{*}{3-stage} & Retained mass (query-wise) & 69.60 & 83.34 & 86.53 & \multirow{3}{*}{--} & \multirow{3}{*}{--} \\
     & Retained mass (block-wise) & 67.56 & 82.04 & 81.00 &  &  \\
     & Readout redundancy & \textit{\underline{82.14}} & 81.10 & 64.14 &  &  \\
    \midrule
    \multirow{3}{*}{4-stage} & Retained mass (query-wise) & 58.74 & 90.72 & 85.58 & 80.16 & \multirow{3}{*}{--} \\
     & Retained mass (block-wise) & 58.55 & 88.04 & 79.60 & 73.17 &  \\
     & Readout redundancy & \textit{\underline{92.87}} & 64.95 & 45.34 & 57.67 &  \\
    \midrule
    \multirow{3}{*}{5-stage} & Retained mass (query-wise) & 69.51 & 94.67 & 84.61 & 89.47 & 79.94 \\
     & Retained mass (block-wise) & 69.21 & 90.70 & 76.03 & 81.29 & 75.14 \\
     & Readout redundancy & \textit{\underline{96.97}} & 71.36 & 28.86 & 50.23 & 34.16 \\
    \bottomrule
  \end{tabular*}
\vspace{-0.4cm}
\end{table*}

Table~\ref{tab:plain_mlp_fusion_4090} shows that fusion reduces latency from $0.333$ to $0.066$ ms at $d=32$ and from $0.750$ to $0.166$ ms at $d=64$, yielding $5.05\times$ and $4.53\times$ speedups, respectively.
Notably, overall MACs remain unchanged.
Peak incremental eager allocation decreases from $128$ to $32$ MiB at $d=32$ and from $256$ to $64$ MiB at $d=64$.

\paragraph{Discussion.}
By avoiding intermediate activation materialization, fusion reduces the idealized logical traffic to
\begin{equation}
    Q_{\mathrm{fused}} = 4(2Nd+4d^2)
    \quad \text{bytes},
\end{equation}
corresponding to an approximately fivefold reduction when $N\gg d$.
The measured $5.05\times$ and $4.53\times$ speedups are consistent with this reduction in activation traffic.
Although other factors, such as kernel scheduling and matrix-multiplication efficiency, may also contribute, these results highlight the memory-bound tendency arising from the materialization of expanded intermediate features in narrow IR FFNs and the importance of fused implementations.

\section{Stage-wise Attention Statistics}
\label{sup_sec:stagewise_statistics}

We analyze a NAFNet~\cite{supp-NAFNet} variant with two SA and two NAF-FFN sublayers per encoder stage and one NAFBlock per decoder stage trained on AIO-5 tasks.
For Table~\ref{tab:attention-mass-design}, we sample 15 images per degradation task using a fixed random seed, yielding 75 images.

\paragraph{Concentration and sharing.}
We measure retained attention mass in three $8\times8$ query regions per SA layer, using a 25\% key budget.
For concentration, each query selects its own highest-weight keys.
For sharing, queries within each $8\times8$ block use a common key set selected by their mean attention probabilities.
Each query retains its own attention weights; only the selected key set is shared.
Results are averaged over queries, heads, layers, and images, with equal weight assigned to each task.
Dense latency is taken from the operator measurements in Table~\ref{tab:stage_attention_latency}, using $512\times512$ inputs on an RTX 4090 at batch size 1; it is measured separately from these attention statistics.
For a fair comparison, we implement dense attention using PyTorch's \texttt{scaled\_dot\_product\_attention} with the FlashAttention~\cite{supp-FA2} backend explicitly enabled.
We also apply K-smoothing in this implementation, as described in Section~\ref{supp:k_smoothing}.

\paragraph{Attention readout redundancy.}
We evaluate all spatial queries and subtract uniform pooling from the attention output:
\begin{equation}
    R_q = G\left(\sum_{k=1}^{N} A_{qk}V_k
                 - \frac{1}{N}\sum_{k=1}^{N}V_k\right),
    \qquad
    \bar R = \frac{1}{N}\sum_{q=1}^{N}R_q,
\end{equation}
where $G$ includes the output projection and learned residual scale, excluding additive bias.
For each SA layer, we compute the spatially shared fraction of this residual energy, then average the two layer-wise ratios with equal weight:
\begin{equation}
    B = \frac{100}{2}\sum_{\ell=1}^{2}
    \frac{\left\langle C^{-1}\|\bar R^{(\ell)}\|_2^2\right\rangle}
         {\left\langle (NC)^{-1}
           \sum_{q=1}^{N}\|R_q^{(\ell)}\|_2^2\right\rangle}.
\end{equation}
Here, $N$ and $C$ denote the number of spatial positions and channels.
$\ell$ indexes the two SA layers in a stage.
For each layer, the brackets average energies over images within each task, then equally over tasks, before taking the ratio.
Each layer receives equal weight in $B$, irrespective of its residual energy.
The residual output combines all heads through $G$; we do not average head-wise ratios.

Full-resolution attention has relatively diffuse weights but a largely spatially shared contribution beyond uniform pooling, with Stage~1 query redundancy reaching 92.87\% under this equal-layer aggregation.
Spatial sharing within an image does not imply input independence or lossless replacement.
We therefore evaluate DSFFN replacement separately in Sec.~\ref{sup_sec:full_resolution}.

\paragraph{Encoder-depth comparison.}
Table~\ref{tab:stagewise_depth} extends the same analysis to three- and five-stage encoders on exactly the same 75 crops.
At the five-stage model's $H/16$ resolution ($16\times16$ tokens), the aligned sampling rule yields two distinct $8\times8$ query regions instead of three.
Across all three encoder depths, readout redundancy is highest at the finest-scale stage, reaching 82.14\%, 92.87\%, and 96.97\% for the three-, four-, and five-stage models, respectively.
Retained attention mass increases from the finest scale to intermediate stages under both query-wise and block-wise routing; block-wise mass peaks at $H/2$ in all three models.

\section{Finest-scale Processing with DSFFNs}
\label{sup_sec:full_resolution}
Motivated by the high attention readout redundancy at full resolution, we compare two alternatives to spatial SA: MDTA~\cite{supp-Restormer}, which leverages globally aggregated, input-dependent channel attention, and our proposed DSFFNs, which use learned, input-independent local convolution kernels.
With an expansion ratio of $\gamma=3$, DSFFN uses an input-dependent gate to modulate the expanded channel responses at each spatial location.
We therefore ask whether this local, channel-wise modulation can yield restoration quality comparable to MDTA at full resolution.
We replace only the two full-resolution encoder SA sublayers, retaining the original H-stage NAF-FFNs and the remaining architecture.
All three models are trained from scratch for 120 pretraining epochs.
All reported checkpoints are evaluated on the same 1,794 AIO-5 inputs, with identical input/target pairs and evaluation settings.

\begin{table}[t]
  \centering
  \caption{Full-resolution SA, MDTA, and DSFFN ($\gamma=3$, $3\times3$) on AIO-5 in PSNR (dB) and SSIM. Denoising: CBSD68, $\sigma=25$. Bold marks unique best results.}
  \label{tab:h_stage_dsffn_aio5}
  \small
  \setlength{\tabcolsep}{3.5pt}
  \renewcommand{\arraystretch}{1.08}
  \resizebox{\columnwidth}{!}{\begin{tabular}{@{}lrrrrrr@{}}
    \toprule
    & \multicolumn{2}{c}{SA baseline} & \multicolumn{2}{c}{MDTA~\cite{supp-Restormer}} & \multicolumn{2}{c}{DSFFN} \\
    \cmidrule(lr){2-3}\cmidrule(lr){4-5}\cmidrule(l){6-7}
    Task / dataset & PSNR & SSIM & PSNR & SSIM & PSNR & SSIM \\
    \midrule
    Haze / SOTS & 28.06 & 0.9669 & 29.46 & \textbf{0.9733} & \textbf{29.68} & 0.9706 \\
    Rain / Rain100L & 36.26 & \textbf{0.9731} & 36.27 & 0.9724 & \textbf{36.37} & 0.9729 \\
    Noise / CBSD68 & \textbf{31.29} & \textbf{0.8901} & 31.28 & 0.8897 & 31.26 & 0.8890 \\
    Blur / GoPro & 29.49 & 0.8864 & \textbf{29.54} & \textbf{0.8865} & 29.47 & 0.8850 \\
    Low-light / LOL & 22.59 & 0.8493 & 22.67 & 0.8514 & \textbf{22.86} & \textbf{0.8547} \\
    \midrule
    Avg. & 29.54 & 0.9132 & 29.84 & \textbf{0.9147} & \textbf{29.93} & 0.9144 \\
    \bottomrule
  \end{tabular}}
\end{table}

Notably, as shown in Table~\ref{tab:h_stage_dsffn_aio5}, DSFFN achieves the highest average PSNR, improving over SA by $0.39$\,dB.
This gain may reflect the benefit of strengthening local context when dense self-attention at lower resolutions already provides global context.
Moreover, given the high readout redundancy across spatial queries, full-resolution SA may add little query-specific information, which could explain why it underperforms DSFFN.
Consistent with this interpretation, adding MDTA to the Stage~1 encoder blocks of HiGLR while retaining their DSFFNs does not improve PSNR on CDD-11 (Table~\ref{tab:stage_operator_ablation}).

\section{More Details about ConvFFN Analysis}
\label{sup_sec:compression_details}

We detail the contribution-wise rank-reduction analysis in
Section~3.4 and Figure~\ref{fig:ffn-width}, which compares the
sensitivity of spatial-interaction and intra-pixel contributions
to output-rank reduction.

\paragraph{Experimental setup.}
We use the official pretrained CDD-11 checkpoints of
BioIR-B~\cite{supp-BioIR} and MoCE-IR-S~\cite{supp-MOCEIR}, with all weights
unchanged.
Each compression is applied simultaneously to all native ConvFFNs~\cite{supp-Restormer}, including refinement blocks and the MoCE-IR latent stage.
PCA statistics are estimated from 220 degraded training images~(20 scenes $\times$ 11 degradation conditions), using three calibration seeds~(20261014, 20261015, and 20261016).
Evaluation uses a disjoint set of 220 test images with fixed $256\times256$ center crops.
We report RGB PSNR averaged equally over degradation conditions and calibration seeds, with matched routing seeds across interventions for MoCE-IR.

\paragraph{Contribution decomposition.}
Let $u$ denote the up-projected features with the up-projection bias removed.
We decompose the depthwise response into a bias-only baseline $B$, an intra-pixel term $L$, and a spatial-interaction term $S$:
\begin{equation}
\begin{aligned}
L_p&=\Bigl(\sum_{\delta}D_{\delta}\Bigr)u_p,\\
S_p&=\sum_{\delta}D_{\delta}(u_{p+\delta}-u_p).
\end{aligned}
\end{equation}
where $D_{\delta}$ is the depthwise kernel at offset $\delta$ and $u$ uses the original zero padding.
Here, $L$ depends on the current feature location, while $S$ captures differences from neighboring locations.
The baseline $B$ includes the depthwise bias and the padded response to the up-projection bias.

Next, let $f$ denote the GELU-product gate followed by the original output projection, including its bias.
We allocate its output using a two-player Shapley decomposition with $B$ fixed, averaging marginal contributions over the two insertion orders~\cite{supp-contextual_decomposition_1801_05453v2}:
\begin{equation}
\begin{aligned}
Y_{\mathrm{spatial}}=\tfrac{1}{2}\bigl[&f(B+S)-f(B)\\
&+f(B+L+S)-f(B+L)\bigr],
\end{aligned}
\end{equation}
and symmetrically for $Y_{\mathrm{intra}}$.
This shares nonlinear interactions symmetrically and gives the exact decomposition
$Y=Y_0+Y_{\mathrm{intra}}+Y_{\mathrm{spatial}}$,
with $Y_0=f(B)$.

\paragraph{Rank reduction.}
We apply mean-preserving response PCA~\cite{supp-zhang_cvpr2015_response_pca} to either contribution $j\in\{\mathrm{intra},\mathrm{spatial}\}$ in the output space, after the original output projection:
\begin{equation}
\widehat{Y}_j=\mu_j+U_{j,k}U_{j,k}^{\top}(Y_j-\mu_j).
\end{equation}
where $\mu_j$ and $U_{j,k}$ are the calibration mean and leading $k$ covariance eigenvectors, fitted separately for each contribution.
We retain $k/C_{\mathrm{out}}\in\{0.25,0.50,0.75\}$ in each layer, keeping the complementary contribution and $Y_0$ exact.

\paragraph{Results and interpretation.}
Table~\ref{tab:supp_convffn_compression} reports absolute PSNR, and Figure~\ref{fig:ffn-width} in the main manuscript shows changes from the uncompressed baseline against output-rank reduction.
Both models exhibit larger losses when the intra-pixel contribution is compressed, with the gap increasing under stronger compression.
These observations motivate decoupling spatial interaction from nonlinear feature expansion, facilitating memory-efficient kernel fusion~\cite{supp-FA2}.

\begin{table}[t]
\centering
\small
\setlength{\tabcolsep}{4pt}
\caption{Contribution-wise PCA compression on CDD-11 crops: mean RGB PSNR (dB) over three calibration seeds.
Rank: retained output-width fraction; 100\%: uncompressed baseline.}
\label{tab:supp_convffn_compression}
\begin{tabular}{lccc}
\hline
Model & Rank & Spatial & Intra-pixel \\
\hline
BioIR-B & 100\% & \multicolumn{2}{c}{30.0905} \\
        & 75\%  & 29.3985 & 27.6374 \\
        & 50\%  & 28.2045 & 25.6020 \\
        & 25\%  & 27.2583 & 21.7189 \\
\hline
MoCE-IR-S & 100\% & \multicolumn{2}{c}{29.3077} \\
          & 75\%  & 29.2484 & 29.1606 \\
          & 50\%  & 29.1262 & 28.4505 \\
          & 25\%  & 28.6028 & 25.9743 \\
\hline
\end{tabular}
\end{table}

\section{Implementation Details}
\label{sup_sec:implementation_details}
We implement our proposed HiGLR with the CUDA~12.6 build of PyTorch~2.14.0 and Triton~3.8.0.

\subsection{Overall Network Architecture}
\begin{figure}[t]
    \centering
    \includegraphics[width=\linewidth]{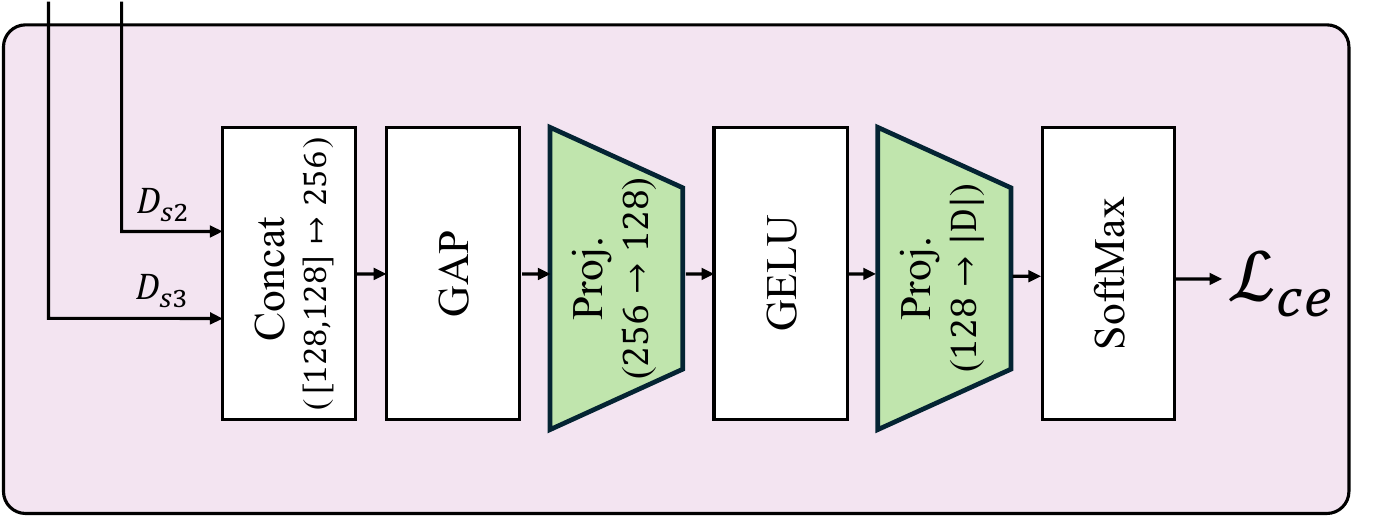}
    \caption{Auxiliary degradation classifier.}
    \label{fig:deg_classifier}
\end{figure}

Table~\ref{tab:architecture} summarizes the encoder and decoder at each resolution.
The encoder traverses stages 1--4, with stage~4 containing only encoder blocks, and the decoder traverses stages 3--1.
Between stages, downsampling uses $2\times2$ convolutions with stride 2, and upsampling uses $1\times1$ convolutions followed by $2\times$ pixel shuffle.
All DSFFNs use expansion ratio $\gamma=3$ and $3\times3$ depthwise convolutions, with spatial widths $R=[16,16,32,64]$ at stages 1--4.
Attention uses 64 channels per head, giving one, two, and four heads at stages 2, 3, and 4, respectively.
Decoder stage 1 uses dense connections, as detailed in Section~\ref{supp:dense_connection}.
At the two intermediate stages, degradation extractors consist of 3$\times$3 convolutions that produce feature maps whose block-wise descriptors condition BRHA.
As shown in Figure~\ref{fig:deg_classifier}, the globally pooled features from both stages are concatenated and passed to an auxiliary classifier trained with cross-entropy supervision.

\begin{table}[t]
  \centering
  \caption{Stage-wise architecture of HiGLR. Feature dimensions are $H_s\times W_s\times C_s$. SA denotes dense self-attention; brackets indicate sequential modules, and $\times n$ denotes repetition. Stages are numbered from finest to coarsest resolution.}
  \label{tab:architecture}
  \small
  \setlength{\tabcolsep}{3pt}
  \renewcommand{\arraystretch}{1.15}
  \begin{tabularx}{\columnwidth}{@{}c >{\centering\arraybackslash}X >{\centering\arraybackslash}X@{}}
    \toprule
    Stage & Encoder & Decoder \\
    \midrule
    1 & $H\times W\times32$ & $H\times W\times32$ \\
      & DSFFN $\times4$ & DSFFN $\times16$ +DenseConnection~\cite{supp-RDN,supp-ETDS} \\
    \midrule
    2 & $\frac{H}{2}\times\frac{W}{2}\times64$ & $\frac{H}{2}\times\frac{W}{2}\times64$ \\
      & [BRHA, DSFFN] $\times4$ & [BRHA, DSFFN] $\times1$ \\
    \midrule
    3 & $\frac{H}{4}\times\frac{W}{4}\times128$ & $\frac{H}{4}\times\frac{W}{4}\times128$ \\
      & [BRHA, DSFFN] $\times4$ & [BRHA, DSFFN] $\times1$ \\
    \midrule
    4 & $\frac{H}{8}\times\frac{W}{8}\times256$ & -- \\
      & [Dense SA, DSFFN] $\times5$ & -- \\
    \bottomrule
  \end{tabularx}
\end{table}

\subsection{K-Smoothing and Numerical Precision}
\label{supp:k_smoothing}
We accelerate dense attention with FlashAttention-2~\cite{supp-FA2}, whose CUDA backend requires FP16/BF16 inputs.
Since low-precision training can degrade restoration quality~\cite{supp-RCANit}, we retain FP32 QKV projections, consistent with prior restoration implementations~\cite{supp-Restormer,supp-PromptIR,supp-HAT,supp-SRFormer,supp-XRestormer}.
When directly casting these projections to BF16 for attention, we observe excessively large query and key norms in certain layers.
Related low-bit attention studies also identify Q/K outliers as a source of numerical sensitivity~\cite{supp-SageAttention,supp-SageBwd}.
Following these studies, we apply K-smoothing.
Technically, we project keys and subtract their spatial mean in FP32 before casting them to BF16.
On CDD-11 ablation, K-smoothing improves PSNR from 30.21 to 30.37\,dB.
In a particularly affected dense attention head, the mean per-token query and key norms are 81.4\% and 52.5\% lower, respectively~(before key mean subtraction), while normalized attention entropy increases from 0.028 to 0.559.

\subsection{Efficient BRHA Implementation}
\label{supp:brha_impl}
Our Triton~\cite{supp-Triton} inference implementation computes routing and coarse aggregation in separate tiled passes without storing the full block-score matrix $S$.
The routing pass merges tile-local top-$k_r$ candidates while accumulating log-sum-exp statistics over all remote blocks.
These statistics normalize the selected log gates over all remote blocks, as required by Eq.~\eqref{eq:bsha_residual}; the query block itself is included separately with a zero log gate.
The resulting block indices and gates are stored for fine attention.
The coarse pass recomputes the score tiles and uses online softmax to accumulate $O_{\coarseidx}$ over all block summaries.
In our RTX 4090 measurements, recomputing the scores yielded slightly lower end-to-end latency and peak allocated memory than storing them.

Fine attention then processes each query block using the selected-block list shared by its queries.
For each selected block, the kernel loads its keys and values, subtracting the block summary from the values to avoid allocating a separate token-resolution residual-value tensor.
The kernel adds the relative position bias and the block log gate to the token scores, and online softmax accumulates the fine output across the selected blocks.
The fine and coarse outputs are then combined according to Eq.~\eqref{eq:bsha_output}.

During the backward pass, fine attention recomputes token-level attention probabilities over the selected blocks using the saved per-query log-sum-exp.
A pass over query blocks computes the query gradients and the gradient contributions to the block log gates and relative position biases.
To compute key and value gradients, we invert the routing map on the GPU by grouping the selected block pairs by destination key/value block.
Within each group, we process the pairs in parallel chunks and sum the partial gradients, avoiding atomic updates to shared key and value gradient entries.

\begin{figure}[t]
    \centering
    \includegraphics[width=\columnwidth]{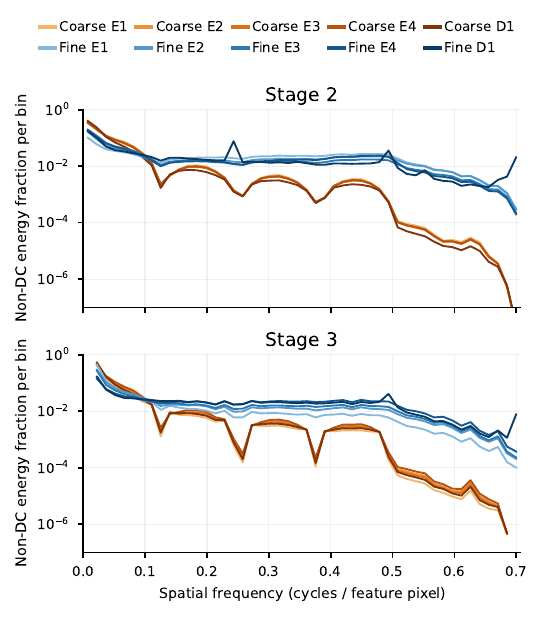}
    \caption{Spatial-frequency distributions of coarse values $U(V_{\coarseidx})$ and fine residuals $V-U(V_{\coarseidx})$ at stages 2 and 3.
    Curves show radially binned power normalized by non-DC energy, averaged over 220 $256\times256$ CDD-11 crops and heads within each layer.
    E1--E4 and D1 denote encoder and decoder blocks, respectively.
    Fine residuals exhibit broader spectra than coarse values, whose blockwise representation favors low frequencies.}
    \label{fig:brha_value_frequency}
\end{figure}

\subsection{Efficient DSFFN Implementation}
\label{supp:dsffn_impl}
Our Triton inference kernels process the expanded channels of DSFFN in chunks.
These kernels use TF32 matrix products with FP32 accumulation and keep intermediate activations in FP32, without BF16 or FP16 casts.
Given the compact spatial response $T$, each chunk computes the corresponding slices of $Z$, $G$, and $W_s(T)$ in Eq.~\eqref{eq:dsffn_output}, applies GELU and sigmoid gating, and immediately accumulates the gated result through $W_o$ into the $C$-channel output.
Thus, no full-image buffers are allocated for $Z$, $G$, $W_s(T)$, or the gated activations.
The kernel scales the residual branch by its learned weight and adds the input features before writing the output.

At lower-resolution stages, the compact projection $W_r$, RMSNorm, and GELU are computed before a separate depthwise convolution, with the $R$-channel intermediate features stored in global memory.
At full resolution, we further reduce memory traffic by fusing this path into the same kernel, using one-pixel halos and recomputing compact features in overlapping tile borders to avoid full-image buffers for both the pre-convolution features and $T$.

During training, we retain the input and compact-path features but do not cache expanded activations for backward.
The backward pass recomputes the expanded pointwise operations from these saved tensors, while reusing the compact projection and depthwise-convolution results.
Recomputed gated activations and value/gate gradients are temporarily materialized for weight-gradient GEMMs.

\subsection{Dense Connection in Decoder 1 Stage}
\label{supp:dense_connection}
\begin{figure}[t]
    \centering
    \includegraphics[width=\linewidth]{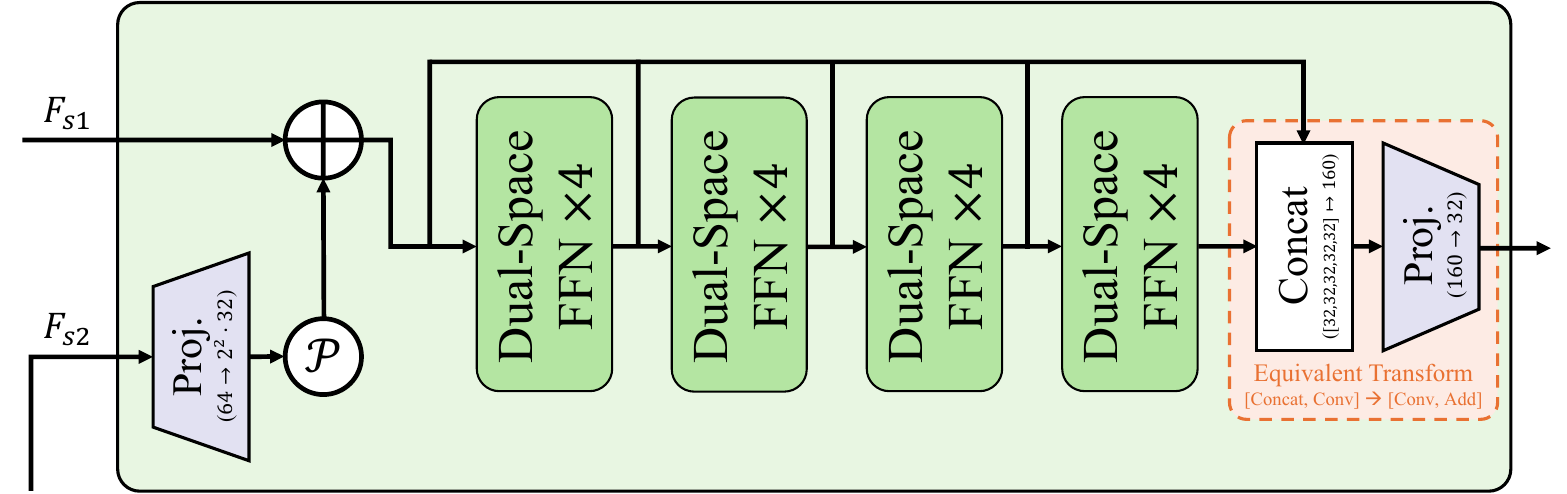}
    \caption{Full-resolution refinement in decoder stage~1.}
    \label{fig:decoder1}
\end{figure}

We empirically find that local refinement at decoder stage~1 is important for restoration quality.
We therefore use 16 DSFFNs at the decoder~1 stage and aggregate intermediate features through dense connections~\cite{supp-RDN}, as illustrated in Figure~\ref{fig:decoder1}.
For dense connections, features before refinement and after every four DSFFNs are fused by a $1\times1$ convolution.
In the CDD-11 ablation, these dense connections improve PSNR by 0.52\,dB (from 29.85\,dB to 30.37\,dB).
Still, retaining and concatenating multiple full-resolution feature maps incurs a substantial memory burden.
We avoid this concatenation through an equivalent transformation~\cite{supp-ETDS}.
Specifically, partitioning the fusion weights along the input-channel dimension gives
\begin{equation}
    W * [F_0;F_4;F_8;F_{12};F_{16}]
    = \sum_{i\in\{0,4,8,12,16\}} W_i * F_i,
\end{equation}
where $F_i$ is the feature after $i$ DSFFNs, $[\cdot;\cdot]$ denotes channel concatenation, and $*$ denotes convolution.
Each feature can thus be projected and accumulated as it becomes available, preserving the fused output while avoiding storage of all five feature maps for fusion at inference.

\begin{table}[t]
\centering
\caption{Effect of FFN expansion on CDD-11 restoration quality and full-model inference cost at $512\times512$ (batch~1, RTX~4090).
Superscripts denote PyTorch ($\S$) and the materialized ($\mathcal{M}$) and recompute ($\mathcal{R}$) Triton~\cite{supp-Triton} implementations. ConvFFN recomputes the branch projections at spatial tile boundaries. ConvNeXt recomputes the normalized depthwise output for each output-channel tile. R, W, and T denote DRAM reads, writes, and their sum per forward pass.}
\label{tab:supp-ffn-full}
\label{tab:supp-ffn-traffic}
\renewcommand{\arraystretch}{1.04}
\footnotesize
\setlength{\tabcolsep}{1.5pt}
\resizebox{\columnwidth}{!}{\begin{tabular}{@{}lrrrrrrrr@{}}
\toprule
Method & $\gamma$ & Params & Latency & Peak & \multicolumn{3}{c}{Traffic (GB)} & PSNR \\
\cmidrule(lr){6-8}
 & & (M) & (ms) & (MiB) & R & W & T & (dB) \\
\midrule
ConvFFN$^{\S}$ & \multirow{3}{*}{1} & \multirow{3}{*}{\textbf{4.03}} & 33.65 & 371.1 & 4.63 & 10.14 & 14.78 & \multirow{3}{*}{29.36} \\
ConvFFN$^{\mathcal{M}}$ &  &  & 24.40 & 307.1 & 3.36 & 5.36 & 8.71 &  \\
ConvFFN$^{\mathcal{R}}$ &  &  & \textbf{20.63} & 283.7 & 0.85 & 3.67 & 4.51 &  \\
\cmidrule(lr){1-9}
ConvFFN$^{\S}$ & \multirow{3}{*}{2} & \multirow{3}{*}{5.43} & 43.02 & 503.5 & 10.43 & 15.80 & 26.23 & \multirow{3}{*}{29.92} \\
ConvFFN$^{\mathcal{M}}$ &  &  & 29.28 & 375.5 & 5.89 & 7.14 & 13.02 &  \\
ConvFFN$^{\mathcal{R}}$ &  &  & 25.75 & 288.2 & 0.80 & 3.66 & \textbf{4.46} &  \\
\cmidrule(lr){1-9}
ConvFFN$^{\S}$ & \multirow{3}{*}{3} & \multirow{3}{*}{6.84} & 54.23 & 636.9 & 16.31 & 20.77 & 37.07 & \multirow{3}{*}{30.12} \\
ConvFFN$^{\mathcal{M}}$ &  &  & 38.51 & 444.9 & 8.44 & 8.73 & 17.17 &  \\
ConvFFN$^{\mathcal{R}}$ &  &  & 29.06 & 293.5 & 0.86 & 3.70 & 4.56 &  \\
\cmidrule(lr){1-9}
ConvFFN$^{\S}$ & \multirow{3}{*}{4} & \multirow{3}{*}{8.24} & 65.56 & 770.3 & 21.35 & 25.74 & 47.09 & \multirow{3}{*}{30.20} \\
ConvFFN$^{\mathcal{M}}$ &  &  & 48.48 & 514.3 & 11.02 & 10.48 & 21.50 &  \\
ConvFFN$^{\mathcal{R}}$ &  &  & 34.50 & 298.9 & 0.82 & 3.70 & 4.52 &  \\
\midrule
ConvNeXt$^{\S}$ & \multirow{3}{*}{1} & \multirow{3}{*}{4.58} & 35.01 & 372.3 & 4.57 & 10.18 & 14.75 & \multirow{3}{*}{29.33} \\
ConvNeXt$^{\mathcal{M}}$ &  &  & 25.38 & 285.0 & 2.56 & 7.33 & 9.90 &  \\
ConvNeXt$^{\mathcal{R}}$ &  &  & 45.96 & 285.0 & 0.85 & 3.68 & 4.53 &  \\
\cmidrule(lr){1-9}
ConvNeXt$^{\S}$ & \multirow{3}{*}{2} & \multirow{3}{*}{6.39} & 40.68 & 507.2 & 8.64 & 13.63 & 22.27 & \multirow{3}{*}{29.46} \\
ConvNeXt$^{\mathcal{M}}$ &  &  & 30.07 & 291.9 & 2.83 & 7.31 & 10.15 &  \\
ConvNeXt$^{\mathcal{R}}$ &  &  & 54.69 & 291.9 & 0.86 & 3.65 & 4.51 &  \\
\cmidrule(lr){1-9}
ConvNeXt$^{\S}$ & \multirow{3}{*}{3} & \multirow{3}{*}{8.20} & 48.34 & 643.0 & 12.17 & 16.96 & 29.13 & \multirow{3}{*}{29.51} \\
ConvNeXt$^{\mathcal{M}}$ &  &  & 33.27 & 299.6 & 2.85 & 7.31 & 10.16 &  \\
ConvNeXt$^{\mathcal{R}}$ &  &  & 67.07 & 299.6 & 0.87 & 3.73 & 4.60 &  \\
\cmidrule(lr){1-9}
ConvNeXt$^{\S}$ & \multirow{3}{*}{4} & \multirow{3}{*}{10.00} & 55.94 & 777.0 & 15.75 & 20.24 & 35.99 & \multirow{3}{*}{29.79} \\
ConvNeXt$^{\mathcal{M}}$ &  &  & 36.37 & 305.6 & 2.86 & 7.34 & 10.20 &  \\
ConvNeXt$^{\mathcal{R}}$ &  &  & 80.49 & 305.6 & 0.87 & 3.65 & 4.52 &  \\
\midrule
DSFFN & 1 & 4.23 & 22.37 & \textbf{283.7} & 0.86 & 3.90 & 4.76 & 30.00 \\
DSFFN & 2 & 5.71 & 22.59 & 289.3 & 0.86 & 3.87 & 4.73 & 30.18 \\
\rowcolor{tablehighlight}
DSFFN (Ours) & 3 & 7.19 & 22.64 & 294.9 & 0.88 & 3.94 & 4.82 & 30.37 \\
DSFFN & 4 & 8.67 & 22.98 & 300.6 & 0.88 & 3.87 & 4.74 & 30.32 \\
\bottomrule
\end{tabular}}
\end{table}

\section{Training Details}
\label{sup_sec:training_details}

\paragraph{Optimization and loss functions.}
We train HiGLR using AdamW with $\beta_1=\beta_2=0.9$,\ a weight decay of $10^{-3}$, and a global batch size of 32.
The learning rate increases linearly from $10^{-6}$ to $10^{-3}$ during warmup and then follows a cosine decay to $10^{-6}$.
The training objective combines reconstruction and degradation classification losses:
\begin{equation}
    \mathcal{L}
    = \mathcal{L}_{1}
    + \lambda_{\mathrm{FFT}}\mathcal{L}_{\mathrm{FFT}}
    + 0.15\mathcal{L}_{\mathrm{MS\text{-}SSIM}}
    + 0.1\mathcal{L}_{\mathrm{CE}},
\end{equation}
where $\mathcal{L}_{\mathrm{MS\text{-}SSIM}}$ denotes
$1-\mathrm{MS\text{-}SSIM}$.
We set $\lambda_{\mathrm{FFT}}=0.1$ during initial training
and reduce it to $0.05$ during AIO fine-tuning.
The classification loss uses label smoothing of 0.1.

\begin{table*}[!b]
\centering
\fontsize{8.5pt}{10pt}\selectfont
\captionsetup{width=0.9\textwidth}
\renewcommand{\arraystretch}{1.12}
\fboxsep0.75pt
\setlength\tabcolsep{1pt}
\caption{Comparison on AIO-3 in PSNR (dB, $\uparrow$) and \colorbox{clblue!50}{SSIM ($\uparrow$)}.
\textbf{Bold} marks the best results (ties included).
$^{\dagger}$Additional frozen CLIP encoder: 88M parameters.}
\label{tab:aio3:higlr}
\begin{tabular*}{0.9\textwidth}{@{\extracolsep{\fill}}l@{\hspace{6pt}}c@{\hspace{8pt}}*{5}{r@{\hspace{3pt}}l@{\hspace{15pt}}}r@{\hspace{3pt}}l@{}}
\toprule
\multirow{2}{*}{Method}
& \multirow{2}{*}{Params.}
& \multicolumn{2}{c}{Dehazing}
& \multicolumn{2}{c}{Deraining}
& \multicolumn{6}{c}{Denoising (CBSD68)}
& \multicolumn{2}{c}{Avg.}\\
\cmidrule(lr){3-4}
\cmidrule(lr){5-6}
\cmidrule(lr){7-12}
& & \multicolumn{2}{c}{SOTS}
& \multicolumn{2}{c}{Rain100L}
& \multicolumn{2}{c}{$\sigma=15$}
& \multicolumn{2}{c}{$\sigma=25$}
& \multicolumn{2}{c}{$\sigma=50$}
& \multicolumn{2}{c}{} \\
\midrule

MPRNet~\cite{supp-MPRNet}
& 16M
& 25.28 & \cc{.955}
& 33.57 & \cc{.954}
& 33.54 & \cc{.927}
& 30.89 & \cc{.880}
& 27.56 & \cc{.779}
& 30.17 & \cc{.899} \\

AirNet~\cite{supp-AirNet}
& 9M
& 27.94 & \cc{.962}
& 34.90 & \cc{.967}
& 33.92 & \cc{.933}
& 31.26 & \cc{.888}
& 28.00 & \cc{.797}
& 31.20 & \cc{.910} \\

PromptIR~\cite{supp-PromptIR}
& 36M
& 30.58 & \cc{.974}
& 36.37 & \cc{.972}
& 33.98 & \cc{.933}
& 31.31 & \cc{.888}
& 28.06 & \cc{.799}
& 32.06 & \cc{.913} \\

VLU-Net~\cite{supp-VLUNet} & 35M$^{\dagger}$ & 30.71 & \cc{.980} & 38.93 & \cc{.984} & 34.13 & \cc{.935} & 31.48 & \cc{.892} & 28.23 & \cc{.804} & 32.70 & \cc{.919} \\

MoCE-IR-S~\cite{supp-MOCEIR}
& 11M
& 30.94 & \cc{.979}
& 38.22 & \cc{.983}
& 34.08 & \cc{.933}
& 31.42 & \cc{.888}
& 28.16 & \cc{.798}
& 32.57 & \cc{.916} \\

MoCE-IR~\cite{supp-MOCEIR}
& 25M
& 31.34 & \cc{.979}
& 38.57 & \cc{.984}
& 34.11 & \cc{.932}
& 31.45 & \cc{.888}
& 28.18 & \cc{.800}
& 32.73 & \cc{.917} \\

BioIR-L~\cite{supp-BioIR}
& 16M
& 31.69 & \cc{.981}
& 38.63 & \cc{.984}
& 34.19 & \cc{\textbf{.937}}
& 31.54 & \cc{\textbf{.895}}
& 28.28 & \cc{\textbf{.809}}
& 32.87 & \cc{.921} \\

AnyIR-S~\cite{supp-AnyIR}
& 9M
& 31.85 & \cc{.982}
& 38.56 & \cc{.983}
& 34.15 & \cc{.936}
& 31.49 & \cc{.893}
& 28.24 & \cc{.806}
& 32.86 & \cc{.920} \\

MIRAGE-T~\cite{supp-MIRAGE} & 6M & 31.81 & \cc{.982} & 38.44 & \cc{.983} & 34.05 & \cc{.935} & 31.40 & \cc{.892} & 28.14 & \cc{.802} & 32.77 & \cc{.919} \\

MIRAGE-S~\cite{supp-MIRAGE} & 10M & 31.86 & \cc{.981} & 38.94 & \cc{\textbf{.985}} & 34.12 & \cc{.935} & 31.46 & \cc{.891} & 28.19 & \cc{.803} & 32.91 & \cc{.919} \\

R2R~\cite{supp-R2R}
& 20M
& 31.40 & \cc{.977}
& 37.46 & \cc{.980}
& 34.10 & \cc{.936}
& 31.45 & \cc{\textbf{.895}}
& 28.22 & \cc{.806}
& 32.53 & \cc{.918} \\

CEA-IR-S~\cite{supp-CEAIR}
& 10M
& 31.90 & \cc{.981}
& 38.62 & \cc{.984}
& 34.16 & \cc{.933}
& 31.51 & \cc{.889}
& 28.27 & \cc{.803}
& 32.89 & \cc{.918} \\

CEA-IR~\cite{supp-CEAIR}
& 22M
& 32.33 & \cc{\textbf{.983}}
& 38.92 & \cc{\textbf{.985}}
& \textbf{34.22} & \cc{\textbf{.937}}
& \textbf{31.58} & \cc{\textbf{.895}}
& 28.32 & \cc{\textbf{.809}}
& 33.07 & \cc{\textbf{.922}} \\

\midrule
HiGLR (\textit{ours}) & 7M & \textbf{32.40} & \cc{.981} & \textbf{39.21} & \cc{\textbf{.985}} & 34.21 & \cc{\textbf{.937}} & 31.57 & \cc{\textbf{.895}} & \textbf{28.33} & \cc{\textbf{.809}} & \textbf{33.14} & \cc{.921} \\
\bottomrule
\end{tabular*}
\end{table*}

\begin{table*}[!b]
\centering
\fontsize{8.5pt}{10pt}\selectfont
\captionsetup{width=0.9\textwidth}
\renewcommand{\arraystretch}{1.12}
\fboxsep0.75pt
\setlength\tabcolsep{1pt}
\caption{Comparison on AIO-5 in PSNR (dB, $\uparrow$) and \colorbox{clblue!50}{SSIM ($\uparrow$)}.
\textbf{Bold} marks the best results (ties included).
$^{\dagger}$Additional frozen CLIP encoder: 88M parameters.}
\label{tab:aio5:higlr}
\begin{tabular*}{0.9\textwidth}{@{\extracolsep{\fill}}l@{\hspace{6pt}}c@{\hspace{8pt}}*{5}{r@{\hspace{3pt}}l@{\hspace{15pt}}}r@{\hspace{3pt}}l@{}}
\toprule
\multirow{2}{*}{Method} & \multirow{2}{*}{Params.} & \multicolumn{2}{c}{Dehazing} & \multicolumn{2}{c}{Deraining} & \multicolumn{2}{c}{Denoising} & \multicolumn{2}{c}{Deblurring} & \multicolumn{2}{c}{Low-light} & \multicolumn{2}{c}{Avg.}\\
\cmidrule(lr){3-4} \cmidrule(lr){5-6} \cmidrule(lr){7-8} \cmidrule(lr){9-10} \cmidrule(lr){11-12}
& & \multicolumn{2}{c}{SOTS} & \multicolumn{2}{c}{Rain100L} & \multicolumn{2}{c}{$\sigma=25$} & \multicolumn{2}{c}{GoPro} & \multicolumn{2}{c}{LOLv1} & \multicolumn{2}{c}{} \\
\midrule
AirNet~\cite{supp-AirNet} & 9M & 21.04 & \cc{.884} & 32.98 & \cc{.951} & 30.91 & \cc{.882} & 24.35 & \cc{.781} & 18.18 & \cc{.735} & 25.49 & \cc{.847} \\

Restormer~\cite{supp-Restormer} & 26M & 24.09 & \cc{.927} & 34.81 & \cc{.962} & 31.49 & \cc{.884} & 27.22 & \cc{.829} & 20.41 & \cc{.806} & 27.60 & \cc{.881} \\

PromptIR~\cite{supp-PromptIR} & 36M & 26.54 & \cc{.949} & 36.37 & \cc{.970} & 31.47 & \cc{.886} & 28.71 & \cc{.881} & 22.68 & \cc{.832} & 29.15 & \cc{.904} \\

VLU-Net~\cite{supp-VLUNet} & 35M$^{\dagger}$ & 30.84 & \cc{.980} & 38.54 & \cc{.982} & 31.43 & \cc{.891} & 27.46 & \cc{.840} & 22.29 & \cc{.833} & 30.11 & \cc{.905} \\

MoCE-IR-S~\cite{supp-MOCEIR} & 11M & 31.33 & \cc{.978} & 37.21 & \cc{.978} & 31.25 & \cc{.884} & 28.90 & \cc{.877} & 21.68 & \cc{.851} & 30.08 & \cc{.913} \\

MoCE-IR~\cite{supp-MOCEIR} & 25M & 30.48 & \cc{.974} & 38.04 & \cc{.982} & 31.34 & \cc{.887} & 30.05 & \cc{.899} & 23.00 & \cc{.852} & 30.58 & \cc{.919} \\

BioIR-L~\cite{supp-BioIR} & 16M & 31.77 & \cc{.981} & 38.75 & \cc{\textbf{.985}} & 31.52 & \cc{\textbf{.894}} & 29.61 & \cc{.889} & 23.29 & \cc{\textbf{.862}} & 30.99 & \cc{.922} \\

AnyIR-S~\cite{supp-AnyIR} & 9M & 31.77 & \cc{\textbf{.982}} & \textbf{39.00} & \cc{.983} & 31.44 & \cc{.892} & 28.52 & \cc{.867} & 23.03 & \cc{.857} & 30.75 & \cc{.916} \\

MIRAGE-T~\cite{supp-MIRAGE} & 6M & 31.35 & \cc{.979} & 38.24 & \cc{.983} & 31.35 & \cc{.891} & 27.98 & \cc{.850} & 23.11 & \cc{.854} & 30.41 & \cc{.912} \\

MIRAGE-S~\cite{supp-MIRAGE} & 10M & 31.45 & \cc{.980} & 38.92 & \cc{\textbf{.985}} & 31.41 & \cc{.892} & 28.10 & \cc{.858} & \textbf{23.59} & \cc{.858} & 30.68 & \cc{.914} \\

R2R~\cite{supp-R2R} & 20M & 30.64 & \cc{.974} & 36.61 & \cc{.975} & 31.35 & \cc{.891} & \textbf{30.93} & \cc{\textbf{.911}} & 22.88 & \cc{.856} & 30.48 & \cc{.921} \\

CEA-IR-S~\cite{supp-CEAIR} & 10M & 31.02 & \cc{.979} & 38.50 & \cc{.984} & 31.45 & \cc{.888} & 29.47 & \cc{.886} & 23.58 & \cc{.858} & 30.80 & \cc{.919} \\

CEA-IR~\cite{supp-CEAIR} & 22M & 31.96 & \cc{.981} & 38.95 & \cc{\textbf{.985}} & \textbf{31.53} & \cc{\textbf{.894}} & 29.90 & \cc{.894} & 23.32 & \cc{.857} & 31.13 & \cc{.922} \\

\midrule
HiGLR (\textit{ours}) & 7M & \textbf{32.17} & \cc{.981} & 38.88 & \cc{.984} & 31.50 & \cc{\textbf{.894}} & \textbf{30.93} & \cc{\textbf{.911}} & 23.02 & \cc{.860} & \textbf{31.30} & \cc{\textbf{.926}} \\
\bottomrule
\end{tabular*}
\end{table*}

\paragraph{Dataset-specific training schedules.}
We largely follow the training protocol of R2R~\cite{supp-R2R}, while adapting the patch sizes and training durations.
To better exploit high-resolution spatial information, we use larger patches in the later training stage and reduce the number of training epochs accordingly.
Despite using larger $256\times256$ patches, HiGLR achieves lower per-step training latency than BioIR-L and MIRAGE-S, both of which use $128\times128$ patches, as shown in Table~\ref{tab:training_cost}.
For AIO-3 and AIO-5, we first train on $128\times128$ patches for 180 epochs and then fine-tune on $256\times256$ patches with a 40-epoch schedule.
The two stages use separate learning-rate schedules with 15 and 7.5 warmup epochs, respectively.
Fine-tuning initializes the model from the pretrained weights and starts with a fresh optimizer.
We maintain an exponential moving average of the model weights with a decay of 0.999 throughout both stages.
For CDD-11, we train on $256\times256$ patches for 240 epochs, including 15 warmup epochs, with EMA.
Unless otherwise specified, the remote-block budget is $k_r=32$ during training and $k_r=16$ during inference.

\section{Details on FFN comparisons}
\label{supp:ffn-traffic}

We compare DSFFN, ConvFFN~\cite{supp-Restormer}, and ConvNeXt~\cite{supp-ConvNeXt} across expansion ratios $\gamma=1$ to $4$ to examine how FFN width affects restoration quality and how each kernel implementation scales with width.
Specifically, we include ConvNeXt because it applies depthwise convolution before channel expansion, so its spatial mixing width is independent of $\gamma$, as in DSFFN.

\paragraph{FFN implementations.}
ConvFFN applies a $3\times3$ depthwise convolution to each of two projected branches of width $\gamma C$. ConvNeXt applies a $7\times7$ depthwise convolution at the input width $C$, followed by LayerNorm and an MLP with hidden width $2\gamma C$.
For both baselines, we evaluate PyTorch ($\S$) and materialized Triton kernels ($\mathcal{M}$). The latter store the two expanded projections for ConvFFN and the normalized depthwise output for ConvNeXt in global memory.
ConvFFN$^{\mathcal{R}}$ recomputes both pointwise projections where the input regions of adjacent spatial tiles overlap and fuses depthwise convolution, gating, and output projection.
ConvNeXt$^{\mathcal{R}}$ recomputes the depthwise convolution and LayerNorm for each output-channel tile and reuses the result across hidden-channel chunks.
DSFFN uses the fused kernels in Sec.~\ref{supp:dsffn_impl}.

\paragraph{Measurement protocol.}
We benchmark the full model on an RTX~4090 with $512\times512$ inputs, batch size~1, $B=8$, and $k_r=16$.
For each implementation and width, we report the median of 110 forward-pass timings collected after warm-up across two independent processes using \texttt{torch.cuda.Event.elapsed\_time()}.

\noindent In separate forward passes, we measure peak GPU memory with \texttt{torch.cuda.max\_memory\_allocated()}.
In separate profiling runs, we measure DRAM reads and writes with Nsight Compute hardware counters and average them over 22 forward passes.

\paragraph{Quantitative results.}
At $\gamma=3$, DSFFN has the highest reported PSNR in Table~\ref{tab:supp-ffn-full} (30.37\,dB), exceeding ConvFFN by 0.25\,dB and ConvNeXt by 0.86\,dB.
Its full-model latency is 22.64\,ms, compared with 29.06\,ms for ConvFFN$^{\mathcal{R}}$ and 33.27\,ms for ConvNeXt$^{\mathcal{M}}$, the fastest implementations of the respective baselines at this width.

\noindent ConvNeXt$^{\mathcal{R}}$ reduces DRAM traffic from 10.16 to 4.60\,GB per forward pass, but its latency rises to 67.07\,ms.
This latency increase is due in part to repeating the same $7\times7$ depthwise convolution and LayerNorm for every output-channel tile covering a given spatial region.
ConvFFN$^{\mathcal{R}}$ is fastest at $\gamma=1$, but widening from $\gamma=1$ to $4$ increases its latency by 67.2\% and that of ConvNeXt$^{\mathcal{M}}$ by 43.3\%.
Over the same range, DSFFN latency rises by only 2.7\%, from 22.37 to 22.98\,ms.
This small increase in latency is consistent with the design of DSFFN, which keeps spatial mixing at a fixed width as $\gamma$ grows and fuses the expanded pointwise operations directly into the output projection.

\section{Full Quantitative Results}
\label{sup_sec:full_quantitative}
To assess restoration performance across diverse degradation types, we report the full per-task comparisons on AIO-3 and AIO-5 in Tables~\ref{tab:aio3:higlr} and~\ref{tab:aio5:higlr}, respectively.
On AIO-3, HiGLR achieves the highest average PSNR of 33.14\,dB, exceeding CEA-IR by 0.07\,dB.
Moreover, on AIO-5, which additionally includes deblurring and low-light enhancement, HiGLR achieves the highest average PSNR and SSIM of 31.30\,dB and 0.926, respectively.
Together, these results support HiGLR's ability to handle diverse restoration tasks within only 7M parameters, achieving higher average PSNR on both benchmarks than the 22M-parameter CEA-IR.

\section{Additional Ablation Studies}
\label{sup_sec:additional_ablations}
We present additional ablations that complement those in the main paper and further support the design of HiGLR.
We examine the number of stages, the remote-block budget across attention variants, the choice of global operator at each stage, and the role of degradation conditioning and its auxiliary loss.

\subsection{Effect of the Number of Stages}
\begin{table}[t]
  \centering
  \definecolor{stageBRHAtext}{HTML}{AD5425}
  \definecolor{stageDenseSAtext}{HTML}{A03D87}
  \caption{Stage-count ablation on CDD-11. All variants use 35 DSFFN blocks. \textcolor{stageBRHAtext}{B} and \textcolor{stageDenseSAtext}{D} denote BRHA and dense self-attention, respectively; $\times N$ denotes the total number of encoder and decoder blocks at each scale, and -- denotes an absent stage. Latency is measured on an RTX 4090 at $512\times512$ with batch size 1.}
  \label{tab:stage_count_ablation}
  \footnotesize
  \setlength{\tabcolsep}{2pt}
  \renewcommand{\arraystretch}{1.15}
  \begin{tabularx}{\columnwidth}{@{}c*{4}{>{\centering\arraybackslash}X}ccc@{}}
    \toprule
    \multirow{2}{*}{Stages} & \multicolumn{4}{c}{Global operator at each scale} & Params & Latency & PSNR \\
    \cmidrule(lr){2-5}
    & $H/2$ & $H/4$ & $H/8$ & $H/16$ & (M) & (ms)$\downarrow$ & (dB)$\uparrow$ \\
    \midrule
    3 & \textcolor{stageBRHAtext}{B} $\times8$ & \textcolor{stageDenseSAtext}{D} $\times7$ & -- & -- & 2.83 & 25.91 & 29.89 \\
    3 & \textcolor{stageBRHAtext}{B} $\times8$ & \textcolor{stageBRHAtext}{B} $\times7$ & -- & -- & 3.07 & 25.32 & 30.20 \\
    \midrule
    \rowcolor{tablehighlight}
    4 (HiGLR) & \textcolor{stageBRHAtext}{B} $\times5$ & \textcolor{stageBRHAtext}{B} $\times5$ & \textcolor{stageDenseSAtext}{D} $\times5$ & -- & 7.19 & 22.64 & 30.37 \\
    \midrule
    5 & \textcolor{stageBRHAtext}{B} $\times4$ & \textcolor{stageBRHAtext}{B} $\times4$ & \textcolor{stageDenseSAtext}{D} $\times4$ & \textcolor{stageDenseSAtext}{D} $\times3$ & 18.02 & 23.07 & 27.18 \\
    5 & \textcolor{stageBRHAtext}{B} $\times4$ & \textcolor{stageBRHAtext}{B} $\times4$ & \textcolor{stageBRHAtext}{B} $\times4$ & \textcolor{stageDenseSAtext}{D} $\times3$ & 18.57 & 25.62 & 28.97 \\
    5 & \textcolor{stageBRHAtext}{B} $\times4$ & \textcolor{stageBRHAtext}{B} $\times4$ & \textcolor{stageBRHAtext}{B} $\times4$ & \textcolor{stageBRHAtext}{B} $\times3$ & 19.53 & 27.67 & 25.71 \\
    \bottomrule
  \end{tabularx}
\end{table}

\begin{table}[t]
\centering
\caption{Stage-wise global operator ablation on CDD-11. All settings use DSFFN for local mixing. -- denotes no global operator.}
\label{tab:stage_operator_ablation}
\setlength{\tabcolsep}{4pt}
\renewcommand{\arraystretch}{1.1}
\footnotesize
\begin{tabular}{@{}lccc@{}}
\toprule
Stage & Stage 1 MDTA & Stage 4 BRHA & \textbf{HiGLR~(Ours)} \\
\midrule
1 & MDTA & -- & -- \\
2 & BRHA & BRHA & BRHA \\
3 & BRHA & BRHA & BRHA \\
4 & Dense SA & BRHA & Dense SA \\
\midrule
PSNR (dB)$\uparrow$ & 30.26 & 29.92 & \textbf{30.37} \\
\bottomrule
\end{tabular}
\end{table}

\begin{figure}[t]
  \input{fig/attention_k_sweep_figure}
\end{figure}

HiGLR switches from BRHA to dense self-attention at $H/8$, the coarsest scale in its four-stage hierarchy.
At this resolution, the cost of attending to all keys is low enough that dense attention is faster than BRHA on $512\times512$ inputs, whereas BRHA is still faster at $H/4$ (Table~\ref{tab:stage_attention_latency}).
To test whether this choice also favors restoration quality, we distribute the same 15 attention blocks across three, four, or five stages and vary the scale at which dense attention takes over.
As shown in Table~\ref{tab:stage_count_ablation}, our four-stage design achieves the highest PSNR and the lowest full-model latency on CDD-11, supporting this choice.

\setcounter{topnumber}{1}
Moreover, the attention statistics in Section~\ref{sup_sec:stagewise_statistics} suggest why BRHA can remain effective even at the bottleneck of a shallower model.
In the three-stage encoder, neighboring queries at $H/4$ can share a small set of keys and still retain most of their attention mass (Table~\ref{tab:stagewise_depth}).
Consistent with this observation, the three-stage restoration model achieves higher PSNR with BRHA at $H/4$ than with dense attention there.
In the five-stage model, extending BRHA all the way to $H/16$ lowers PSNR, suggesting that access to keys outside the shared set is useful at this coarser scale.
The best five-stage result therefore comes from reserving dense attention for $H/16$.
Switching one stage earlier, at $H/8$, makes the model faster but sacrifices restoration quality.

\begin{table}[t]
\centering
\caption{Inference latency at $k_{r}=16$ (RTX~4090, $512\times512$, batch size 1).}
\label{tab:attention_variant_latency}
\footnotesize
\setlength{\tabcolsep}{2pt}
\begin{tabular}{@{}lcccccc@{}}
\toprule
Variant & Coarse & Sparse & \shortstack{Separate\\$K$/$V$} & $V_r$ & log gate & \shortstack{Latency\\(ms)} \\
\midrule
Sparse-only & -- & $\checkmark$ & -- & -- & $\checkmark$ & 21.47 \\
Log-gate OFF & $\checkmark$ & $\checkmark$ & -- & $\checkmark$ & -- & 22.30 \\
Baseline~(BRHA) & $\checkmark$ & $\checkmark$ & -- & $\checkmark$ & $\checkmark$ & 22.64 \\
\bottomrule
\end{tabular}
\end{table}

\subsection{Stage-wise Global Mixers}
With the four-stage hierarchy fixed, we ask whether a global mixer would complement the local DSFFNs at full resolution and whether BRHA could replace dense attention at the bottleneck (Table~\ref{tab:stage_operator_ablation}).
At full resolution, the residual readout of dense spatial attention is largely shared across queries (Table~\ref{tab:stagewise_depth}), suggesting limited benefit from computing a separate spatial readout for every query.
MDTA offers another efficient way to incorporate global context at this scale by mixing channels.
Yet local DSFFNs perform on par with MDTA in the AIO-5 replacement study (Table~\ref{tab:h_stage_dsffn_aio5}), and the CDD-11 variant that adds MDTA to the Stage~1 encoder shows no PSNR gain over HiGLR.
These observations support keeping the finest stage local and relying on coarser stages for global context.

At the bottleneck, we ask whether BRHA can replace dense self-attention without sacrificing restoration quality.
In the four-stage encoder, block-wise routing with a fixed 25\% key budget leaves more attention mass outside the selected set at $H/8$ than at $H/2$ or $H/4$ (Table~\ref{tab:stagewise_depth}).
Consistent with this observation, replacing dense attention with BRHA at $H/8$ lowers CDD-11 PSNR (Table~\ref{tab:stage_operator_ablation}). Extending BRHA to the $H/16$ bottleneck of the five-stage model has the same effect (Table~\ref{tab:stage_count_ablation}).
Also note that, at $H/8$, the token count is small enough that dense attention runs faster per module than BRHA on $512\times512$ inputs (Table~\ref{tab:stage_attention_latency}).
Thus, keeping dense attention at the four-stage bottleneck favors both restoration quality and inference speed at that scale.

\subsection{Comparisons on Attention Variants}
\label{sec:attention_ablation_protocol}
Reducing the remote-block budget $k_r$ limits the regions that fine attention can sample, while coarse attention still aggregates information from the entire image.
To test whether this global context helps preserve restoration quality at small budgets, we vary $k_r$ at inference on CDD-11 with each model's weights fixed.
We compare Sparse-only, which attends directly to values in the selected blocks, with two hybrid variants that combine dense attention over pooled block values with sparse attention over residual token values.
The two hybrid variants differ in whether the routing scores also gate fine attention. In full BRHA, this log gate allows the restoration loss to update the scores used for block selection during training.

\begin{figure*}[!t]
  \begin{minipage}[t]{\columnwidth}
    \vspace{0pt}
    \captionsetup{type=figure}
    \input{fig/classifier_input_umap_figure}
  \end{minipage}\hfill
  \begin{minipage}[t]{\columnwidth}
    \vspace{0pt}
    \captionsetup{type=figure}
    \input{fig/four_scene_selections_figure}
  \end{minipage}
\end{figure*}

When the budget falls from 16 remote blocks to one, Log-gate OFF loses less PSNR than Sparse-only, and full BRHA loses less than half as much as Sparse-only (Figure~\ref{fig:attention_k_sweep}~(a)).
This ordering suggests that the coarse path preserves context from blocks omitted by fine attention and that learning the routing scores through the log gate further improves robustness to a smaller budget.
The advantage also holds in absolute PSNR. Log-gate OFF with six remote blocks and full BRHA with just one both outperform Sparse-only attending to approximately 25\% of the key/value blocks (Figure~\ref{fig:attention_k_sweep}~(b)).

Notably, the coarse path can cover all blocks efficiently because it reuses the routing scores to weight the pooled values.
At $k_r=16$, measured full-model latency only increases by 0.83\,ms from Sparse-only to Log-gate OFF and by a further 0.34\,ms from Log-gate OFF to full BRHA (Table~\ref{tab:attention_variant_latency}).
Sharing the same scores across routing, coarse aggregation, and fine-attention gating thus lets BRHA retain restoration quality at small budgets with modest inference overhead.

\setcounter{topnumber}{99}
\subsection{Degradation Embedding and CE Loss}
To assess the need for degradation-based conditioning in BRHA, we conduct ablation studies and visualize the learned degradation representations to examine whether they capture degradation-specific information.
On CDD-11, removing both degradation conditioning and the CE loss yields 30.17\,dB PSNR.
Retaining degradation conditioning without CE gives 30.25\,dB, while using both achieves 30.37\,dB.
These gains suggest that conditioning block selection on the input degradation helps BRHA identify regions that are informative for restoration.
Moreover, the UMAP~\cite{supp-UMAP} visualization in Figure~\ref{fig:classifier-input-umap} shows clear separation of features across degradation conditions, indicating that the model extracts degradation-specific representations.
Finally, Figure~\ref{fig:four-scene-selections} shows that top-$k_r$ block selection patterns vary substantially across degradation conditions for the same scene and fixed central query, illustrating that our BRHA adaptively selects relevant blocks according to the degradation.

\begin{figure*}[!t]
  \centering
  \includegraphics[width=\textwidth]{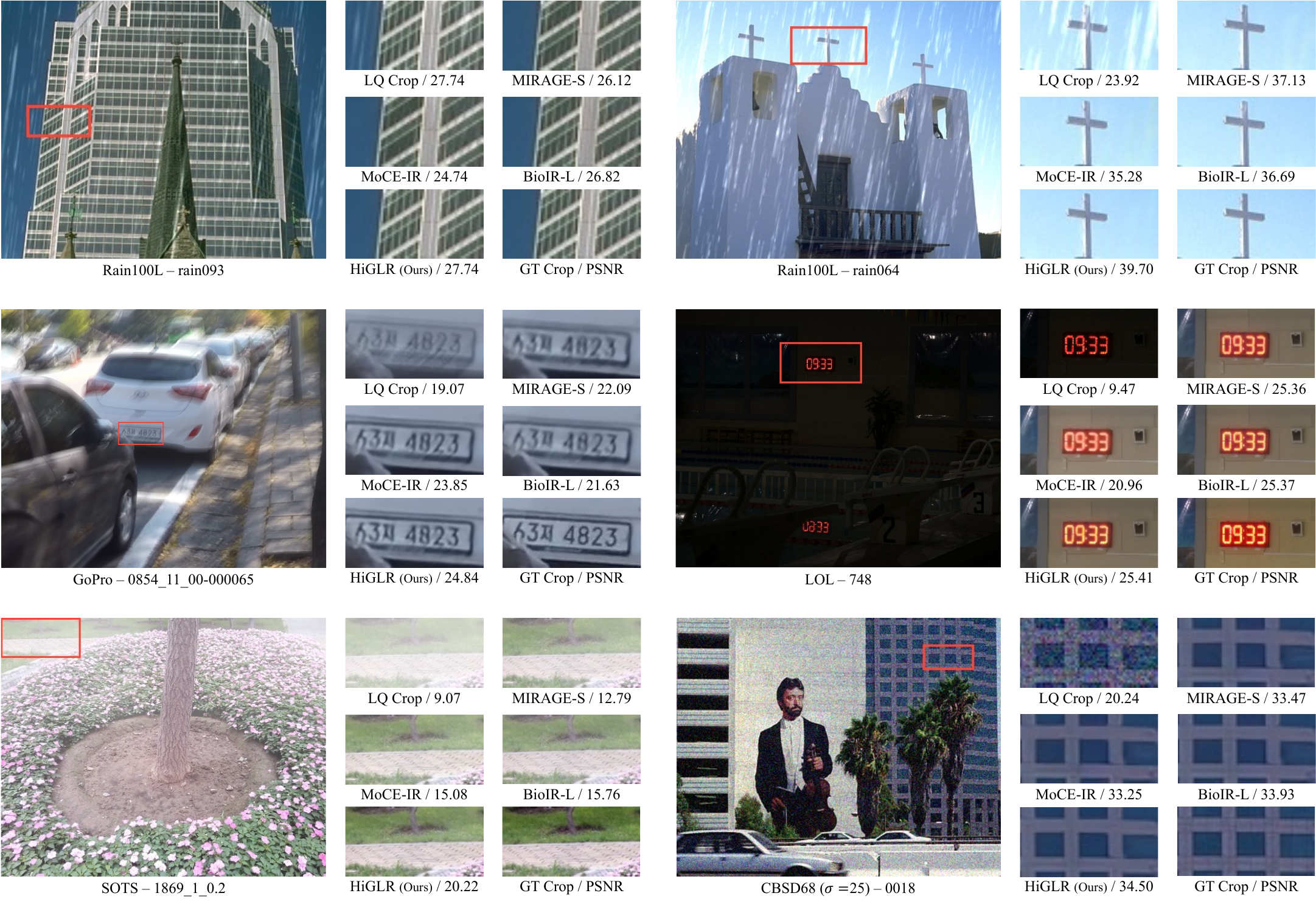}
  \caption{Visual comparisons for deraining, deblurring, low-light enhancement, dehazing, and denoising. Red boxes mark the regions shown in the enlarged crops. Numbers denote PSNR in dB.}
  \label{fig:supp_visual_results}
\end{figure*}

\section{Comparisons on Training Cost}
We leverage $256\times256$ patches for CDD-11 training and AIO fine-tuning to provide self-attention with a wider spatial context (Section~\ref{sup_sec:training_details}).
Despite the larger patch size, HiGLR maintains high training efficiency through our memory-efficient implementation, which recomputes fine-attention probabilities and expanded DSFFN activations during the backward pass (Secs.~\ref{supp:brha_impl} and~\ref{supp:dsffn_impl}).
For example, as shown in Table~\ref{tab:training_cost}, HiGLR requires only 234.24\,ms per training step and 4.37\,GB of peak allocated GPU memory on an RTX~4090 with a batch size of four, whereas BioIR-L, MIRAGE-S, and MoCE-IR require 250.65--344.16\,ms and 7.11--15.35\,GB, despite using $128\times128$ patches with four times fewer pixels.
These results demonstrate that HiGLR can leverage a wider spatial context while achieving lower training time and memory consumption than BioIR-L, MIRAGE-S, and MoCE-IR.

\begin{table}[t]
    \centering
    \caption{Training cost comparisons~(RTX 4090, batch 4): time per step and peak allocated GPU memory.}
    \label{tab:training_cost}
    \small
    \setlength{\tabcolsep}{4pt}
    \renewcommand{\arraystretch}{1.12}
    \begin{tabular}{@{}lcrr@{}}
        \toprule
        Method & Patch & Latency (ms) & Memory (GB) \\
        \midrule
        R2R
            & $224^2$ & 116.54 & 3.33 \\
        \rowcolor{tablehighlight}
        HiGLR (Ours)
            & $256^2$ & 234.24 & 4.37 \\
        BioIR-L
            & $128^2$ & 250.65 & 15.35 \\
        MIRAGE-S
            & $128^2$ & 277.24 & 7.11 \\
        MoCE-IR
            & $128^2$ & 344.16 & 9.39 \\
        \bottomrule
    \end{tabular}
\end{table}

\begin{table}[t]
\centering
\captionsetup{skip=4pt}
\caption{Inference latency (ms)$\downarrow$ across GPUs and input resolutions at batch size~1.}
\label{tab:supp-latency}
\footnotesize
\setlength{\tabcolsep}{3.25pt}
\renewcommand{\arraystretch}{1.0}
\begin{tabular}{@{}lrrrrrr@{}}
\toprule
& \multicolumn{2}{c}{RTX~4090} & \multicolumn{2}{c}{H200} & \multicolumn{2}{c}{RTX~PRO~6000} \\
\cmidrule(lr){2-3}\cmidrule(lr){4-5}\cmidrule(l){6-7}
Method & $512^2$ & $1024^2$ & $512^2$ & $1024^2$ & $512^2$ & $1024^2$ \\
\midrule
MoCE-IR-S~\cite{supp-MOCEIR} & 76.44 & 342.34 & 45.41 & 150.64 & 49.78 & 221.39 \\
BioIR-B~\cite{supp-BioIR} & 89.01 & 412.69 & 50.88 & 185.73 & 62.26 & 294.67 \\
MIRAGE-S~\cite{supp-MIRAGE} & 83.48 & 327.05 & 45.68 & 163.75 & 50.65 & 226.13 \\
R2R~\cite{supp-R2R} & 31.76 & 58.14 & 13.40 & 31.78 & 18.50 & 37.66 \\
CEA-IR-S~\cite{supp-CEAIR} & 96.07 & 445.19 & 62.88 & 250.28 & 61.46 & 288.14 \\
\midrule
\rowcolor{tablehighlight}
HiGLR~(Ours) & 22.64 & 94.58 & 17.32 & 65.76 & 16.91 & 73.47 \\
\bottomrule
\end{tabular}
\end{table}

\section{Latency across Resolutions and GPUs}
\label{sup_sec:efficiency_resolutions_gpus}
We extend the inference benchmark in Table~\ref{tab:efficiency} to three GPUs and two input resolutions (Table~\ref{tab:supp-latency}).
Notably, our HiGLR ranks first or second in latency among the six methods in every setting.
At $512\times512$, HiGLR is fastest on the RTX~4090 (22.64\,ms) and RTX~PRO~6000 (16.91\,ms), and ranks second on the H200 with a latency of 17.32\,ms.
At $1024\times1024$, HiGLR ranks second on all three GPUs, taking 65.76--94.58\,ms per image, or 56--71\% less time than the third-fastest method on each GPU.
These results demonstrate that HiGLR consistently achieves high inference efficiency across diverse GPU architectures and input resolutions, highlighting its practical scalability.

\section{More Visual Results on AIO-5}
\label{sup_sec:more_visual_results}
While Figure~\ref{fig:main_visual} in the main manuscript demonstrates the strong qualitative performance of HiGLR under complex compound degradations, its visual fidelity under single-degradation scenarios remains to be verified.
To this end, we conduct additional qualitative comparisons across the five restoration tasks in AIO-5, as shown in Figure~\ref{fig:supp_visual_results}.
Across these diverse tasks, HiGLR consistently recovers fine-grained structural details while yielding degradation-robust reconstructions.
Taken together with the CDD-11 comparisons, these results confirm that HiGLR generalizes reliably across both isolated and compound degradations, preserving faithful scene details regardless of degradation complexity.

\end{document}